\RequirePackage{fix-cm}

\documentclass[smallextended]{svjour3}   

\smartqed
\usepackage{graphicx}
\usepackage{float}
\graphicspath{{images/}}
\usepackage{caption}
\usepackage{subcaption}
\usepackage{setspace}

\usepackage{mathptmx}  
\usepackage{amsmath,bm}
\usepackage{amssymb,mathtools}

\usepackage{algpseudocode}
\usepackage{algorithm}
\usepackage{xcolor}
\usepackage{caption}
\usepackage[title]{appendix}

\makeatletter
\newcommand{\new}[1]{{\textcolor{blue}{#1}}}

\makeatletter
\newcommand\remembertext[2]{
  \immediate\write\@auxout{\unexpanded{\global\long\@namedef{mytext@#1}{#2}}}%
  #2%
}

\newcommand\recalltext[1]{%
  \new{\ifcsname mytext@#1\endcsname
    \@nameuse{mytext@#1}%
  \else
    ``??''
  \fi
}}
\makeatother

\journalname{Cognitive Computation}

\begin{document}

\title{A developmental approach for training deep belief networks\thanks{Conflict of interest: The authors declare that they have no conflict of interest.}}

\titlerunning{Developmental approach for deep belief networks}

\author{Matteo Zambra         \and
        Alberto Testolin      \and
        Marco Zorzi
}

\institute{Matteo Zambra \at
               Department of General Psychology,  University of Padova, Via Venezia 8, 35131 Padova, Italy\\ (now at Department of Electric and Mathematical Engineering -- IMT Atlantique, Brest, France)\\ \vspace{-3mm}
           \and
           Alberto Testolin* \at
              Department of General Psychology, University of Padova, Via Venezia 8, 35131 Padova, Italy and
              Department of Mathematics, University of Padova, Via Trieste, 63, 35121 Padova, Italy\\\vspace{-3mm}
           \and
           Marco Zorzi* \at
              Department of General Psychology and Padova Neuroscience Center, University of Padova, Via Venezia 8, 35131 Padova, Italy and
              IRCCS San Camillo Hospital, via Alberoni 70, Venice Lido, Italy \\\vspace{-3mm}
           \and
           * Correspondence: alberto.testolin@unipd.it, marco.zorzi@unipd.it
}

\date{}
\maketitle

\vspace{-2cm}

\begin{abstract}
\textit{Background:} Deep belief networks (DBNs) are stochastic neural networks that can extract rich internal representations of the environment from the sensory data. DBNs had a catalytic effect in triggering the deep learning revolution, demonstrating for the very first time the feasibility of unsupervised learning in networks with many layers of hidden neurons. These hierarchical architectures incorporate plausible biological and cognitive properties, making them particularly appealing as computational models of human perception and cognition. However, learning in DBNs is usually carried out in a greedy, layer-wise fashion, which does not allow to simulate the holistic maturation of cortical circuits and prevents from modeling cognitive development.
\textit{Method:} Here we present \emph{iDBN}, an iterative learning algorithm for DBNs that allows to jointly update the connection weights across all layers of the model. We evaluate the proposed iterative algorithm on two different sets of visual stimuli, measuring the generative capabilities of the learned model and its potential to support supervised downstream tasks. We also track network development in terms of graph theoretical properties and investigate the potential extension of iDBN to continual learning scenarios.
\textit{Results:} DBNs trained using our iterative approach achieve a final performance comparable to that of the greedy counterparts, at the same time allowing to accurately analyze the gradual development of internal representations in the deep network and the progressive improvement in task performance.
\textit{Conclusions:} Our work paves the way to the use of \emph{iDBN} for modeling neurocognitive development.

\end{abstract}

\section{Introduction}
\label{intro}
Despite the fact that the most popular approach for training deep neural networks is based on supervised learning \cite{lecun2015}, the first demonstration of the potential of deep learning stemmed from the discovery of efficient unsupervised learning methods for stochastic neural networks known as Deep Belief Networks (DBNs) \cite{hinton2006fast,hinton2006reducing}. Since their introduction, DBNs have been successfully used in many challenging tasks, ranging from computer vision \cite{lee2009convolutional} to acoustic modeling \cite{mohamed2011acoustic}, traffic flow prediction \cite{huang2014deep} and breast cancer classification \cite{abdel2016breast}. These energy-based models have some unique properties compared to other unsupervised deep learning approaches, such as the ability to represent compositional structure \cite{du2019implicit,tubiana2017emergence} and the possibility to be interpreted in terms of well-established theoretical principles rooted in statistical physics \cite{melko2019restricted}.

The capability of learning deep generative models using Hebbian-like mechanisms also makes DBNs particularly relevant for cognitive modeling research \cite{zorzi2013modeling}. Indeed, this class of models offers a principled account for the functional role of top-down processing supported by feedback loops, at the same time providing a bridge to higher-level descriptions of cognition in terms of Bayesian computations \cite{friston2010,testolin2016probabilistic}. \remembertext{Comment-1.3.1}{Compared to shallow generative models, hierarchical generative networks allow to study the emergence of increasingly more complex representations of the sensory signal, thus allowing to simulate a wide range of high-level perceptual and cognitive functions, such as numerosity perception \cite{stoianov2012,zorzi2018emergentist,testolin2020visual}, letter perception \cite{testolin2017letter,sadeghi2017learning}, orthographic processing \cite{di2013deep}, development of object-action associations \cite{grzyb2019children} and the appearance of visual hallucinations caused by damage in cortical areas \cite{reichert2013charles}.} Sparse variants of DBNs have also been used to simulate physiological properties of neurons in the primary and secondary visual cortex \cite{lee2008sparse}. Notably, inference algorithms for DBNs can be implemented using biologically realistic sampling schemes, which explain unique aspects of low-level brain dynamics \cite{buesing2011neural} and can be efficiently reproduced in spiking models \cite{o2013real}. Finally, hierarchical generative models like the DBN offer important insights into the functional role of spontaneous brain activity, both in terms of top-down predictive signals during task execution and in terms of generative replays during rest \cite{pezzulo2021}.

\remembertext{Comment-1.2}{However, learning in DBNs has traditionally relied on a greedy, layer-wise training approach: the connection weights of layer $n$ are changed only after the layer $n - 1$ has been fully trained. Moreover, the trained weights are frozen and do not further change while learning takes place at higher layers. Although efficient from a computational perspective, this learning modality is clearly implausible from a biological standpoint. Indeed, the greedy approach implies that information is not passed to any higher-order network until learning at the lower level can be stopped because it has reached a (somewhat arbitrary) criterion. Though brain development shows variability across regions and may peak at different times, synaptogenesis begins at about the same time in distant regions such as the visual and the prefrontal cortex \cite{huttenlocher1997}. Moreover, a substantial degree of plasticity is preserved in adulthood even in the visual cortex (see \cite{castaldi2020}, for review). Finally, spontaneous brain activity, which is thought to be a manifestation of top-down dynamics of generative models \cite{pezzulo2021}, is already structured into distinct cortical networks at birth  \cite{fransson2007}.}

\remembertext{Comment-1.3.2}{Besides these neurobiological considerations, another key limitation of the greedy training approach is that it makes the DBN unsuitable for modeling human development. Neural network models are particularly attractive for understanding developmental phenomena \cite{elman1996} because the trajectories in task performance or in the emergence of internal representations can be examined during learning and compared to human empirical data. Moreover, learning trajectories in neural network models can be analysed as a function of initial starting conditions to study the emergence of developmental disorders \cite{zorzitestolin2022}. In the cognitive modeling literature, however, simulations based on DBNs have focused on adult performance (i.e., fully-trained models) and developmental investigations based on unsupervised learning have adopted alternative learning paradigms based on deep autoencoders \cite{testolin2020numerosity} trained with error backpropagation \cite{rumelhart1986} to circumvent this problem.}

In this work we propose a novel learning scheme for tuning the entire hierarchy of connections in a DBN iteratively (hereafter, iDBN), using a variant of the original Contrastive Divergence (CD) learning algorithm \cite{hinton2002}. \remembertext{Comment-1.1}{Through extensive simulations on two different sets of visual stimuli, we demonstrate that the proposed approach can achieve the same final accuracy of the greedy counterpart, at the same time allowing for a precise tracking of the developmental trajectory of the models. We further show that an alternative developmental scheme based on full-stack propagation of top-down information does not converge to an optimal solution, suggesting that recurrent processing between adjacent layers is a key ingredient to successfully drive learning through local signals.} In a first set of simulations we rely on the popular MNIST data set of handwritten digits \cite{lecun1998gradient}, with the goal of validating the proposed iterative scheme on a well-known benchmark. \remembertext{Comment-1.4.1}{In this setup, we also show that our developmental learning scheme can be extended to continual learning scenarios \cite{parisi2019continual}, where the model exploits interleaved learning to incorporate knowledge from another domain (in our case, handwritten letters) without incurring in catastrophic forgetting \cite{mccloskey1989catastrophic,french1999catastrophic}.} Furthermore, we carry out graph theoretical analyses to investigate how structural properties of the network gradually emerge during learning. We finally consider a more recent data set consisting of images containing a variable number of items, which has been used to investigate the perception of visual numerosity in humans and machines \cite{stoianov2012,testolin2020visual}, to demonstrate how the iDBN can be used to simulate developmental trajectories in the acquisition of cognitive skills.

\section{Methods}
\label{sec:methods}
In this section we will briefly review the theoretical foundations of deep belief networks and describe their classical, greedy learning algorithm. We will then introduce our iterative learning approach and describe the materials and methods used in our simulations.

\subsection{Deep Belief Networks}
The building block of a DBN is the Restricted Boltzmann Machine (RBM \cite{ackley1985learning}; for a recent overview see \cite{zhang2018overview}), which is a bipartite network composed by two separate sets of neurons: \emph{visible} neurons, which constitute the interface with the sensory environment and are thus usually clamped to the input data, and \emph{hidden} (equivalently called `latent') neurons that allow to capture high-order correlations in the data distribution. Learning in RBMs consists in discovering a set of latent features that can be used to compactly describe the statistical regularities in the data distribution, by creating an internal model of the environment that can be used to generate plausible activation patterns in the visible neurons. The lack of connections between neurons in the same layer makes it easy to compute the data-dependent and model-dependent statistics used in the Constrastive Divergence (CD) algorithm \cite{hinton2002}, because units in the same layer are conditionally independent given the activation of the other layer. Deep belief networks are created by stacking together several RBMs \cite{hinton2006fast} (see Figure 1a), thus allowing to exploit hierarchical composition of the features learned by the individual RBMs \cite{hinton2007}.

\begin{figure}[]
    \captionsetup{justification=centering}
    \centering
    \begin{subfigure}[]{0.9\textwidth}
        \includegraphics[width=\textwidth]{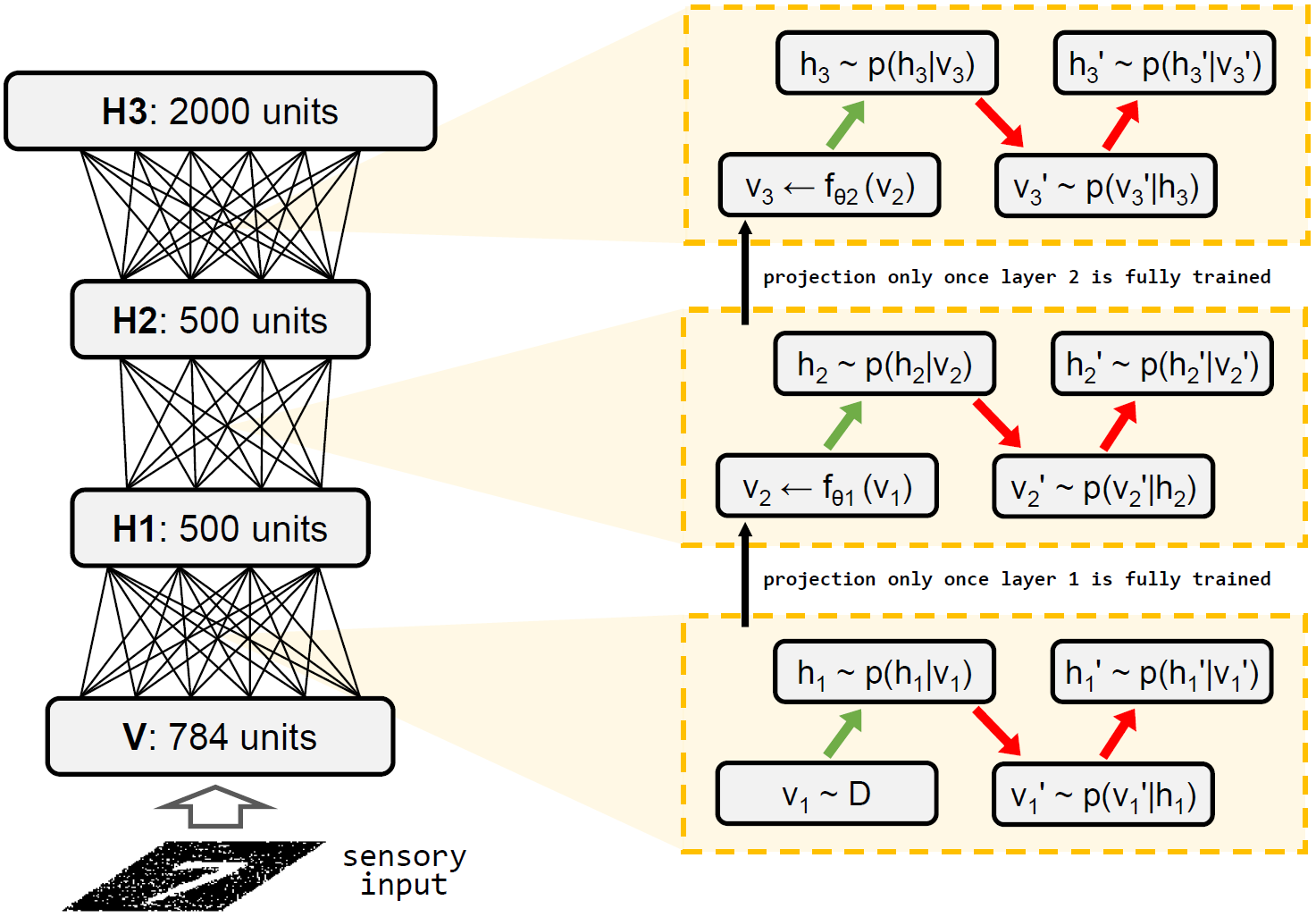}
        \caption{Schematic illustration of the DBN architecture and the greedy, layer-wise learning scheme.}
        \label{fig:greedy_scheme}
    \end{subfigure}
    
    \begin{subfigure}[]{0.43\textwidth}
        \includegraphics[width=\textwidth]{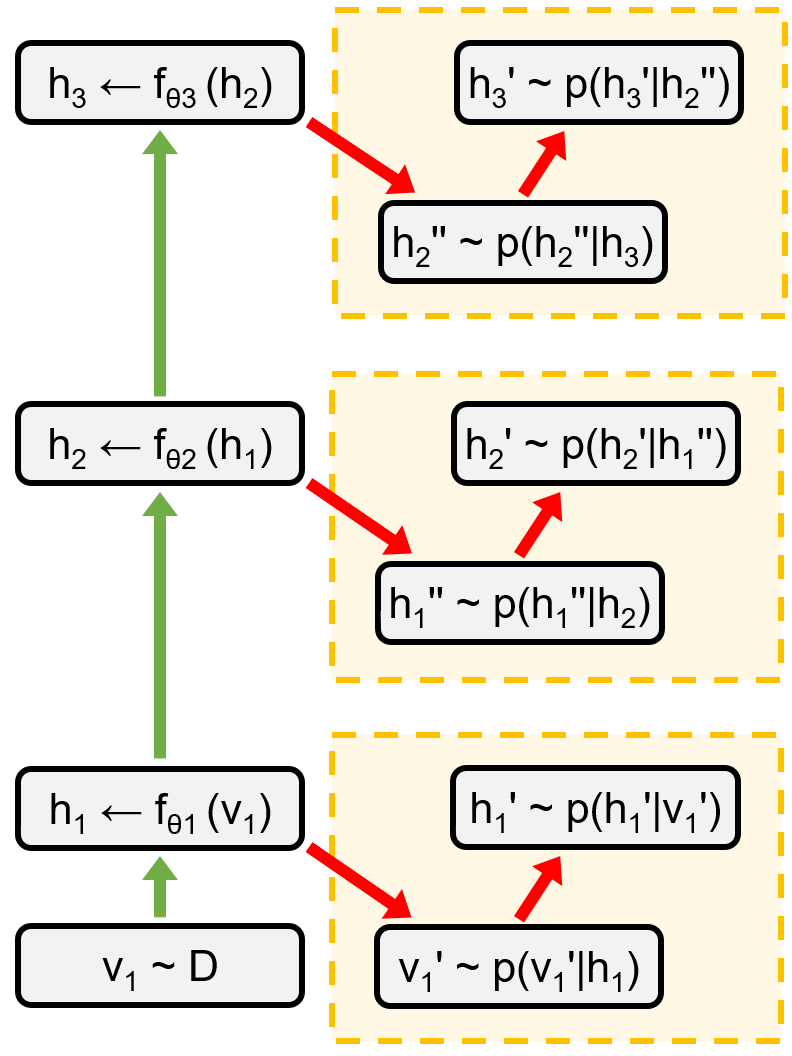}
        \caption{Iterative learning scheme.}
        \label{fig:iterative_scheme}
    \end{subfigure}
    \hspace{8mm}
    \begin{subfigure}[]{0.43\textwidth}
        \includegraphics[width=\textwidth]{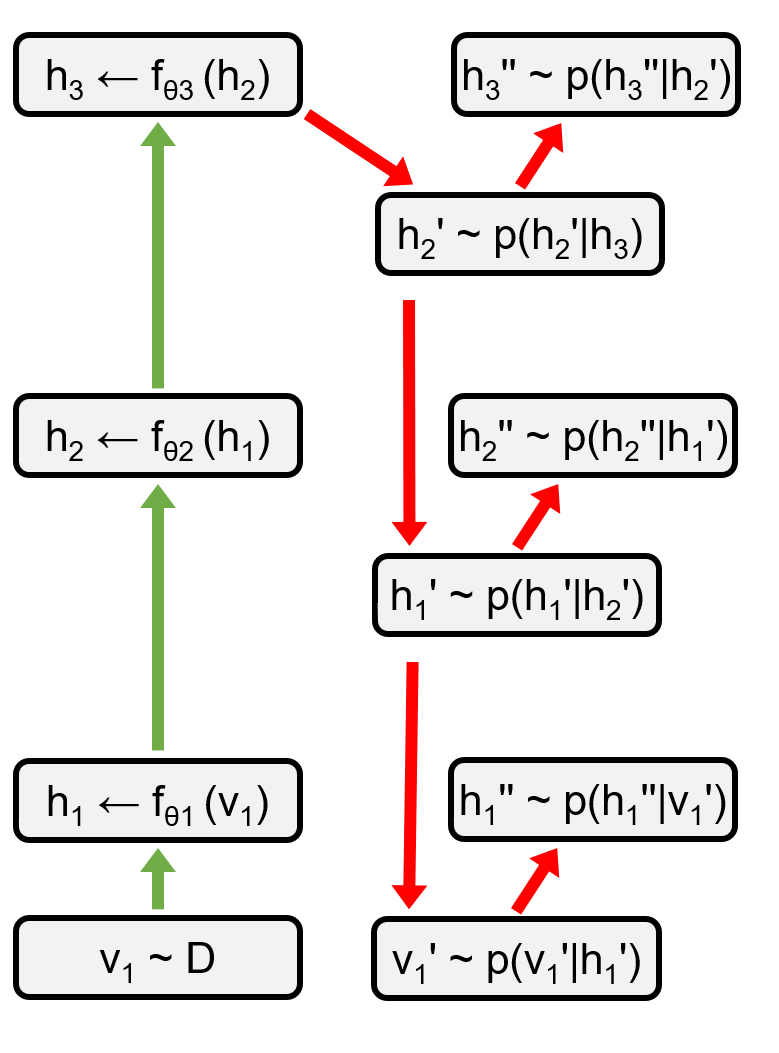}
        \caption{Full-stack learning scheme.}
        \label{fig:fullstack_scheme}
    \end{subfigure}
    \captionsetup{justification=justified}
    \caption{Graphical representation of the architecture of a 3-layer Deep Belief Network and the learning schemes implemented in the present work. Green arrows represent bottom-up recognition connections, while red arrows represent top-down generative processing. Yellow boxes enclose local computations. We consider the case of CD1, CD-$k$ can be recovered by repeating the sampling steps $k$ times. $v \sim D$ identifies a data instance sampled from the training set and $\bm{h}_i$ represents the hidden activities of layer $i$. In the greedy scheme (a) hidden layers are trained sequentially, from bottom to top, and input signals are never projected into layer \emph{l} unless learning at layer \emph{l - 1} is completed. In the iterative scheme (b) input signals are immediately propagated through the entire deep network, and top-down processing is performed locally at each layer to jointly learn all connection weights. In the full-stack scheme (c) both feed-forward propagation and top-down processing occur over the entire deep network.}
    \label{fig:dbn_learning}
\end{figure}

\subsubsection{Greedy layer-wise learning}
For the sake of our argument, it is useful to make an explicit distinction between bottom-up \emph{recognition} connections and top-down \emph{generative} processing (represented by green and red arrows, respectively, in Figure~\ref{fig:dbn_learning}). During the recognition (also called ``inference'') phase, the sensory pattern is clamped on the visible neurons, and the hidden neurons are activated in a bottom-up fashion in order to infer the most likely configuration (i.e., activation pattern over the hidden neurons) that could have produced the observed data. During the generation phase, the visible neurons are not clamped to the data, and all neurons are activated in a top-down fashion in order to produce a plausible activation pattern, that is, to generate a sample from the internal model.

In the classical approach \cite{hinton2006fast}, the RBM constituting the bottom layer of the DBN is initially trained using the input data. Once this first layer has been \emph{fully} trained, the weights of the first RBM are ``frozen'' and the input data set is projected into the activation space of the hidden neurons, thus creating a new training set that is used as input for the second RBM  (black arrows in Figure~\ref{fig:greedy_scheme}). Once training of the second layer is completed, the data projection is made likewise to create the training set for the third layer; the procedure is repeated for all the layers constituting the DBN, as shown in Figure~\ref{fig:greedy_scheme}. Note that the data patterns projected into the deeper layers are created by computing the conditional probability distribution of the hidden neurons given the activation of the neurons observed in the layer below \cite{hinton2012}.

\subsubsection{Iterative joint learning}
In contrast to the greedy approach, the proposed iDBN learning algorithm attempts to update at once all the parameters of the hierarchical generative model, regardless of the depth of the respective layer (see Figure~\ref{fig:iterative_scheme}).

As noted before, in the greedy algorithm deeper layers are trained sequentially, using as input the projection of the data through the weights resulting from learning in the previous layers. In our novel iterative algorithm, instead, the training patterns for the deeper layers are immediately created following each sensory experience, by propagating the input across the entire processing hierarchy (green arrows in Figure~\ref{fig:iterative_scheme}). This process mimics the \emph{fast feed-forward sweep} observed in cortical circuits, where neuronal activity is rapidly routed to a large number of visual areas after stimulus presentation \cite{lamme2000distinct,vanrullen2007power}. It should be noted that in our algorithm the complete feed-forward sweep occurs even during the initial learning phase, where all the connection weights are randomly initialized, which makes learning of the subsequent layers challenging since the data distribution (i.e., hidden unit activation) is non-stationary. Concurrently with the fast feed-forward sweep, top-down generative connections are locally used to reconstruct the data representations at each level of the hierarchy (red arrows in Figure~\ref{fig:iterative_scheme}), mimicking the kind of processing supported by recurrent and horizontal connections within cortical areas \cite{lamme2000distinct,kreiman2020beyond}\footnote{It should be noted that, following the initial feed-forward sweep, the generative phases at subsequent levels of the hierarchy still occur sequentially, because the model-driven activation of hidden neurons at layer $t$ should not interfere with model-driven activation of the same neurons at layer $t+1$. Other families of generative models, such as the Deep Boltzmann Machine \cite{salakhutdinov2009deep}, resolve this potential interference by incorporating top-down influences during learning, but nevertheless require a greedy training strategy.}.
A schematic pseudo-code that illustrates the implementation details of our iterative algorithm is reported below\footnote{The complete source code is freely available to download:\\ https://github.com/MatteoZambra/Developmental-Approach-DBN}.

\begin{algorithm}
\small
	\begin{algorithmic}[1]
		\caption{Iterative learning in iDBN}
		
		\State Define model architecture and initialize the model parameters $\bm{\theta} = \{W_i, \bm{a}, \bm{b}_i\}$, $W_i$ are the weights matrices for layers $i = 1, \dots, N_L$, $\bm{a}$ is the bias of the visible neurons, $\bm{b}_i$ are the hidden neurons biases for the layers $i = 1, \dots, N_L$. For recommendations on a good choice of the parameters please refer to Reference~\cite{hinton2012}
		\State Prepare the training set $\bm{X}_0 \leftarrow Data$
		\While {not convergence}
			\For {layer $i = 1, \dots, N_L$}
				\State Store the input activations in a temporary variable $\bm{X}_{\text{tmp}} \leftarrow \bm{X}_i$
				\For {each $\bm{v}_i$ in $\bm{X}_i$}
				    \State $\bm{h}_i \sim p(\bm{h}_i | \bm{v}_i) = \sigma(W_i\,\bm{v}_i + \bm{b}_i)$, data-driven hidden activations
				    \State $\bm{v}'_i \sim p(\bm{v}'_i | \bm{h}_i) = \sigma(W_i^{t}\,\bm{h}_i + \bm{a}_i)$, model-driven visible activations
				    \State $\bm{h}'_i \sim p(\bm{h}'_i | \bm{v}'_i) = \sigma(W_i\,\bm{v}'_i + \bm{b}_i)$, model-driven hidden activations
				    \State $\Delta W^{+} = \langle \bm{v}_i \cdot \bm{h}_i \rangle$ and $\Delta W^{-} = \langle \bm{v}'_i \cdot \bm{h}'_i \rangle$
				    \State $\Delta \bm{a}^{+} = \langle \bm{v}_i \rangle$ and $\Delta \bm{a}^{-} = \langle \bm{v}'_i \rangle$
				    \State $\Delta \bm{b}^{+} = \langle \bm{b}_i \rangle$ and $\Delta \bm{b}^{-} = \langle \bm{h}'_i \rangle$
				    \State $\Delta \bm{\theta}_i = \Delta \bm{\theta}^{+} - \Delta \bm{\theta}^{-}$, with $\bm{\theta} = \{W, \bm{a}, \bm{b}\}$
				    \State $\bm{\theta}_i = \text{StochasticGradientDescent}(\bm{\theta}_i, \Delta \bm{\theta}_i)$
				\EndFor
				\State Reset the input data to $\bm{X}_i \leftarrow \bm{X}_{\text{tmp}}$
			\EndFor
		\EndWhile
	\label{alg:iterative}
	\end{algorithmic}
\end{algorithm}

\subsubsection{Full-stack joint learning}
\label{sec:fullstack}
An alternative way to jointly learn all weights of a DBN could be to first propagate the input across the entire processing hierarchy (as in the first phase of iDBN) and then produce top-down reconstructions starting from the deepest layer back to the sensory layer (see Figure~\ref{fig:fullstack_scheme}). This processing scheme is simpler to implement, but suffers from the vanishing gradient problem encountered in standard deep learning settings \cite{bengio1994}. We use this full-stack developmental scheme as a benchmark for our iterative developmental scheme.

\subsection{Simulations}
\subsubsection{MNIST data set}
The same DBN structure is used and held fixed during the simulations for both greedy and iterative learning. In order to allow for a fair comparison, we maintained the same architecture and hyper-parameters of the original model \cite{hinton2006fast}, which was composed of one visible layer with 784 neurons, two hidden layers with 500 neurons each and a final hidden layer with 2000 neurons (see Figure~\ref{fig:greedy_scheme}). Connection weights are initialized with random values sampled from a Normal distribution $N(0, 0.01)$, while biases are intialized to zero. We also tested an alternative initialization scheme known as the Glorot initialization \cite{glorot2010}, in order to evaluate learning convergence under more advanced weight initialization strategies. The weights matrices initialized with the Glorot scheme are multiplied by a factor of $0.1$ to make them compatible with the range observed in the Normal initialization. The learning rate is set to $\lambda = 0.01$ and the weight decay coefficient is set to $\alpha = 0.0001$. The momentum parameter $\nu$ is set to $0.5$ in the initial learning stage and then updated to $0.9$ after 5 epochs of training. \remembertext{Comment-2.6.1}{The loss is minimized using standard stochastic gradient descent, implemented through CD1 learning.} The model is trained for 50 epochs. Further tests, using both Normal and Glorot initialization schemes, also included dropout regularization \cite{srivastava2012} with the probability of unit presence set to $p=0.1$.

The first quantity inspected to evaluate the quality of the learned models is the accuracy of a linear readout at each hidden layer: since in DBNs the input patterns are non-linearly transformed from one layer to the next one, the internal representations are supposed to become more linearly separable as we move up in the hierarchy \cite{zorzi2013modeling}. The accuracy of a simple Ridge classifier is used as a measure of separability.

The performance of the trained models is then assessed in three image generation tasks: 1) reproduction of clean images, 2) completion of partially occluded images and 3) denoising of images corrupted by noise. In the second case, an arbitrary number of subsequent rows in the image matrix are set to zero (i.e., turned to black). In the latter case, all values $\{X_{ij}\}_{i,j = 1}^{28}$ (being 28 the number of side-pixel of the images) are spoiled by adding Gaussian noise, that is $X_{ij} \leftarrow X_{ij} + \epsilon$, $\epsilon \sim N(0, 0.5)$. Images are fed to the visible layer of the DBN, propagated through all its hidden layers and then fed back to the visible layer through feedback connections. In this way the noisy / corrupted samples can be adjusted according to the internal model learned by the DBN. \remembertext{Comment-2.6.2}{Original samples and the corresponding reconstructions are quantitatively compared using mean squared error (hereafter MSE) between input patterns and reconstruction. For each case, the error is computed as $\| X^{0} - X^{\text{r}} \|$, where the superscript $0$ denotes the original sample and $r$ means that the image has been reconstructed (reproduced, recreated or denoised) and $\| \cdot \|$ denotes an $L^2$ norm.} Model performance is averaged over 10 model runs with different random initialization in order to assess the robustness of the analyses. We also visualize the receptive fields of neurons at different hidden layers to qualitatively assess the type of features (i.e., internal representations) learned by the DBNs.

\subsubsection{Continual learning}

To further support the cognitive validity of our approach, we investigate whether the proposed developmental algorithm could effectively deal with a challenging continual learning scenario, where the DBN should learn a generative model of data distributions provided incrementally, without forgetting knowledge obtained during the preceding stages. In such scenario it is well-known that neural networks suffer from catastrophic forgetting, whereby knowledge learned during the subsequent stages completely disrupts previously learned information (for review, see \cite{french1999catastrophic}). \remembertext{Comment-1.4.2}{Several solutions have been proposed to mitigate this issue, and here we specifically consider one that can be readily implemented within our framework: \emph{interleaved learning} \cite{complementarySystems2016}. We consider a somewhat simplified version of interleaved learning, where unsupervised learning during a second training phase takes advantage of both patterns from the new target distribution and patterns belonging to the previous data set. More advanced implementations could exploit deep generative replay \cite{generative2017} to directly sample previously-learned patterns from the hierarchical generative model, allowing to build the mixed dataset in a more data-efficient way.}

\remembertext{Comment-1.4.3}{Following the first stage of unsupervised learning on handwritten digits, the DBN is exposed to a subset of handwritten letters from the EMNIST data set \cite{emnist2017}. In one setup the DBN is trained sequentially, which means that only letter patterns are used to train the network during the second stage. In the interleaved setup, instead, during the second stage the unsupervised training set includes both digits and letters. In order to balance the number of patterns and output classes, we randomly sample 20,000 digits from the MNIST data set and we evenly sample 20,000 uppercase letters from the first 10 EMNIST classes. Continual learning performance is probed by monitoring both digit recognition accuracy (using the read-out classifier trained during the first learning stage) and letter recognition accuracy (training a new read-out classifier on the EMNIST patterns).}

\subsubsection{Graph analysis}
\label{sec:resultsgraph}
Recent work has shown that graph theory can be successfully used to study deep learning models from a network science perspective \cite{testolin2019,zambra2020}. In order to investigate how topological properties of the graph derived from the DBN might emerge during the course of unsupervised learning, we thus performed a graph analysis on the structure of the deep network during learning over the MNIST data set.

From the trained DBN we extract the weights matrices and define a graph having the same architecture and connections weights as the DBN. A methodological difficulty is posed by the continuity of the connection weights: in order to determine the nodes degrees, we thus prune the network by \emph{binarizing} the connections according to a suitably chosen cut-off threshold. \remembertext{Comment-2.5}{
A sensible choice of the cut-off threshold is a value that allows to remove redundant connections and to keep those that contribute most to the signal propagation through the network. Note that the concept of ``redundant'' and ``important'' connections is largely arbitrary and not obvious to assess. For the sake of the structural analysis, however, this choice is based on the numerical magnitude of the connections strengths. A range of such thresholds has been set to $\{0.2, 0.4, 0.6, 0.8, 1, 1.25, 1.5\}$, and structural analysis has been performed for each of these values.} A cut-off threshold of $c$ discards the weights in the interval $[-c, c]$, while those outside the interval are kept for the network analysis.

As customary in network science, the structural properties of the ``real'' networks (DBNs) are compared with the same properties observed in a random counterpart. 
\remembertext{Comment-2.4}{A random ``replica'' of the network is generated according to its architectural characteristics (such an number of nodes) and its local properties (e.g., mean node degree or edge probability). The process is composed of two main steps: 1) generate a random replica of the real network and 2) perform the structural analyses on both the instances. In our case, the real network was compared with an analogous binomial graph generated using the probability of edge existence, computed as the ratio between number of effective edges and the maximum number of potential edges (which depends of the number of node couples). This random replica is generated in such a way to have the same architecture of the DBN, in particular it is a stack of bipartite binomial graphs with the same number of nodes as the DBN layers.} Due to this architecture, the node connectivity is constrained by the number of nearby layers. For example, one node of the first layer is not allowed to be connected with one node of the third layer or superior. \remembertext{Comment-2.3}{Such graph structure poses a problem in the characterization of the nodes degree distribution, which is used in network science to evaluate whether a graph is random or derived from a real network. These latter (might them be natural, biological, technological or social networks) typically have a degree distribution well described by a power-law \cite{barabasi1999}, while random graphs have a degree distribution that depends on the method they are generated with. For example, binomial random graphs have a degree distribution that follows, by design, the Binomial distribution.}
In our setup, the constrained nodes connectivity would lead the normalization of the degree distribution to have such architectural bias. To mitigate this effect, we chose to weight the degree of each node according to its potential maximum degree, as described in detail in the Appendix~\ref{app:graphanalysis}.

\subsubsection{Numerosity data set}
\label{sec:numds}
The ``numerosity'' data set was first introduced by Stoianov and Zorzi \cite{stoianov2012}, who demonstrated that the approximate number of objects in a visual scene can be estimated by a hierarchical processing architecture that learns to extract increasingly more abstract representations from the sensory input in a completely unsupervised way (sample images are provided in Figure~\ref{fig:sz}). Here we focus on the development of ``number acuity'' in the network, which has been recently investigated using a developmental approach based on deep autoencoders \cite{testolin2020numerosity}. Number acuity can be measured using a numerosity discrimination task, where the network is asked to classify any image in terms of containing a larger or smaller number of objects with respect to a given reference number. Also in this case, to ensure a fair comparison with the original model~\cite{stoianov2012}, we considered the same model architecture and task.
The DBN is composed by a visible layer composed of $900$ visible neurons, while the first and the second hidden layers have $80$ and $400$ neurons, respectively. We adopted a Normal initialization scheme, where weights are initialized with random values sampled from $N(0,0.1)$. The learning rate and weight decay are set to $\lambda = 0.1$ and $\alpha = 0.0002$, respectively, and the initial and final momentum are set to $\nu = 0.5$ and $0.9$, again with a momentum switch at epoch $5$. The model is trained for 100 epochs.

To assess the number acuity of the network (also known as ``internal Weber fraction''), a linear classifier is applied to the deepest layer of the model, with the goal of establishing whether the hidden neurons' activation correspond to an input image containing a numerosity larger than a reference number (i.e., $8$ or $16$). This task becomes trivial in the limit of the difference between the given numerosity $n_i$ and the reference number $N_{\text{ref}}$ being large. For example, it is easier to tell which numerosity is smaller among $4$ and $16$, but it is harder for $15$ and $16$. For this reason, each reference numerosity has an associated window of numerosities used for the comparison: $\{5,\dots, 12\}$ for $N_{\text{ref}} = 8$ and $\{10,\dots, 24\}$ for $N_{\text{ref}} = 16$, so that the ratios $r_i = n_i / N_{\text{ref}}$ yield the range $[0.65, 1.25]$.
The percentage of correct classifications is analysed as a function of these numerical ratios: each value $r_i$ has associated the percentage of correct classifications $y_i$. This ensemble of points $(r, y)$ is used to fit a psychometric function corresponding to a logistic curve, defined by 

\begin{equation}\label{eq:cdf}
    y = 1 - \Phi(\mu = r, \sigma = 2\,w)\lvert_{x = 0}
\end{equation}
being $\Phi$ the cumulative distribution function of the Normal distribution and $w$ the Weber fraction.

Our interest is to compare the progressive refinement of the Weber fraction during the unsupervised learning phase. The final performance can be directly compared with the Weber fraction achieved by a DBN trained using the greedy scheme \cite{stoianov2012}, while the developmental trajectories can be compared to the learning curves recently reported for deep autoencoders \cite{testolin2020numerosity}.

\section{Results}
\label{sec:results}
\subsection{MNIST data set}
\label{sec:mnist_results}

\subsubsection{Readout accuracy and reconstruction error trends}
\label{ssec:rc}
Results discussed in this section only refer to the DBN configuration with Normal initialization and no dropout, since it turned out that differences with Glorot initialization and inclusion of dropout are negligible. The reader may refer to appendix~\ref{app:robustness} for the complete results.

\begin{figure}[t]
    \centering
    
    \begin{subfigure}[b]{\textwidth}
        \includegraphics[width=\textwidth]{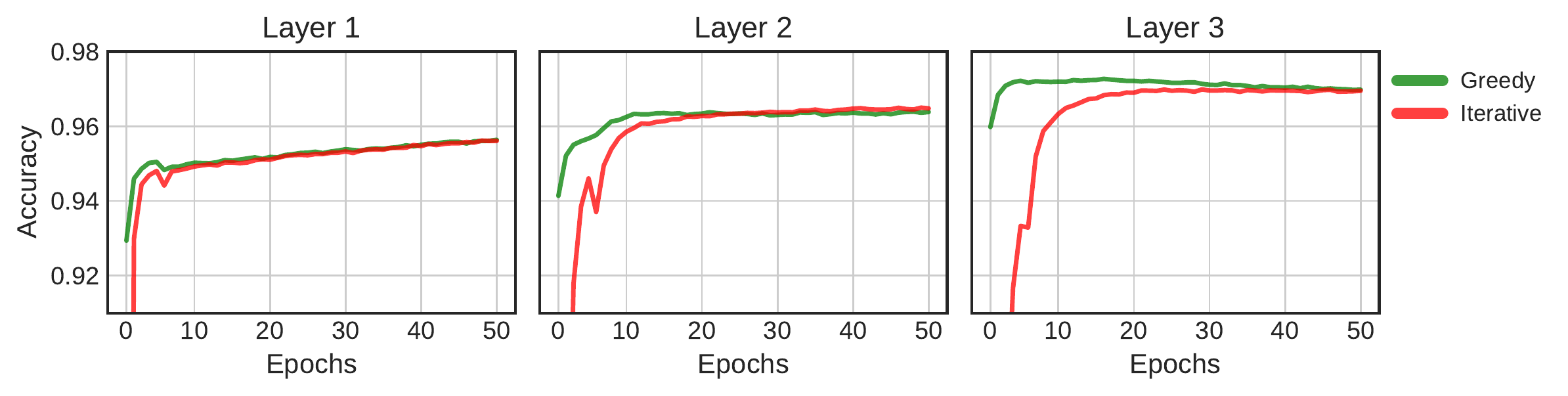}
        \caption{Readout profiles at different layers of the DBN hierarchy during training.}
        \label{fig:cd1_readout_normal}
    \end{subfigure}
    
    \begin{subfigure}[b]{\textwidth}
        \includegraphics[width=\textwidth]{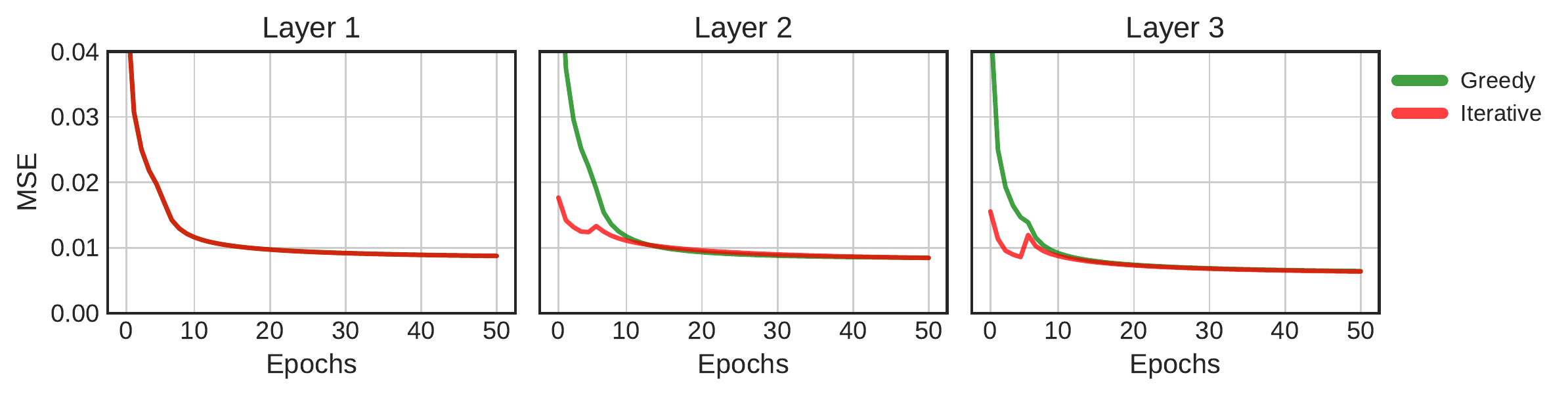}
        \caption{Reconstruction errors at different layers of the DBN hierarchy during training.}
        \label{fig:cd1_cost_normal}
    \end{subfigure}
    
    \begin{subfigure}[b]{0.85\textwidth}
        \includegraphics[width=\textwidth]{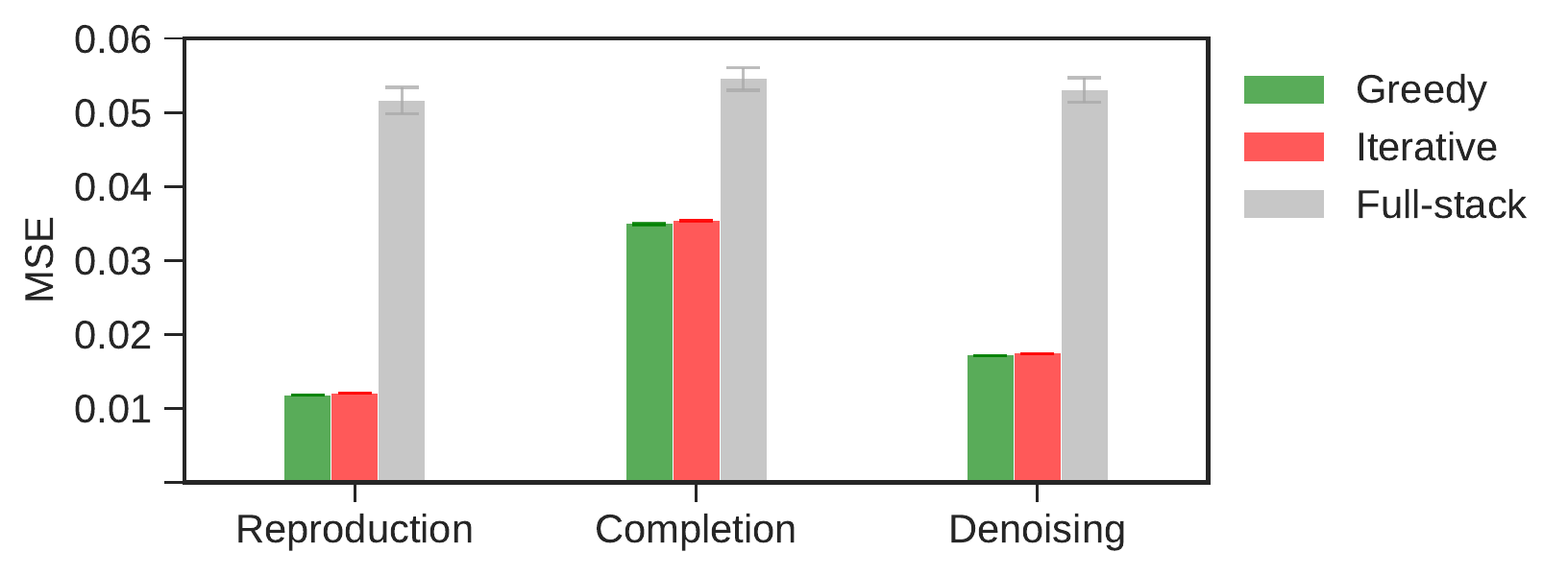}
        \caption{Average error on the image generation tasks (error bars represent standard error).}
        \label{fig:cd1_mses_normal}
    \end{subfigure}
    \caption{Performance of the greedy vs. iterative learning schemes during learning (top and middle panels) and at the end of the unsupervised learning phase (bottom panel). For the latter case we also report the generation capabilities of the alternative developmental scheme based on full-stack propagation.}
    \label{fig:learning-curves-normal}
\end{figure}

Figure~\ref{fig:cd1_readout_normal} shows that the readout accuracy increases with depth, suggesting that during the course of unsupervised learning the internal representations became more disentangled (i.e., linearly separable). The reconstruction errors (MSE at each hidden layer) keep decreasing monotonically and eventually converge (Figure~\ref{fig:cd1_cost_normal}). Not surprisingly, results related to the first layer are almost perfectly overlapping for the greedy and iterative learning schemes: the activation of the visible layer corresponds to the raw data in both cases, thus the learned weights do not depend on the training modality. For the second and third layers, instead, the greedy scheme achieves higher readout accuracy in fewer epochs: however, this effect is due to the fact that the previous layers have been already completely trained, thus providing a head-start for the upper layers. At the end of training, the final accuracy is indistinguishable. Notably, the alternative developmental variant based on full-stack propagation does not exhibit the same optimal convergence, as highlighted by the poor reconstruction error measured in the first hidden layer (see Figure~\ref{fig:iterative_vs_fullstack}).

\subsubsection{Generative capabilities}
As shown in Figure~\ref{fig:cd1_mses_normal}, at the end of the learning phase the greedy and iterative DBNs achieve equivalent generative capabilities (samples of generated images are provided in Figure~\ref{fig:data-reconstruction}). The completion task appears as the more challenging, probably because when entire regions of the images are corrupted it is difficult to generate plausible completions. The full-stack developmental version does not converge to a satisfactory generative model, as highlighted by its poor capability in all generation tasks.

\subsubsection{Emergent internal representations}
Receptive fields are useful to qualitatively inspect what kind of features are learned by the different layers of the hierarchical model during training. Such inspection is done by visualizing the connection weights of a given neuron in the input space \cite{zorzi2013modeling}. The first hidden layer is easy to inspect, since we just need to plot the weights matrix of the first layer $W_1$. When it comes to neurons of the second hidden layer it is necessary to compose the weights matrices, so that it is still possible to represent the visualization in terms of the visible layer dimension. The weights matrices are simply multiplied, thus producing a linear combination: we can simply plot some chosen rows of the product $W_2 W_1$ to look at the receptive fields of the second layer neurons, $W_3 W_2 W_1$ for the third layer, and so forth. As shown in Figure~\ref{fig:receptivefields_normal}, the greedy and iterative learning schemes developed qualitatively similar receptive fields across the entire processing hierarchy.

\subsubsection{Resilience to catastrophic interference}
\label{sec:continual_learning}
\begin{figure}[t]
    \centering
    \includegraphics[width=0.8\textwidth]{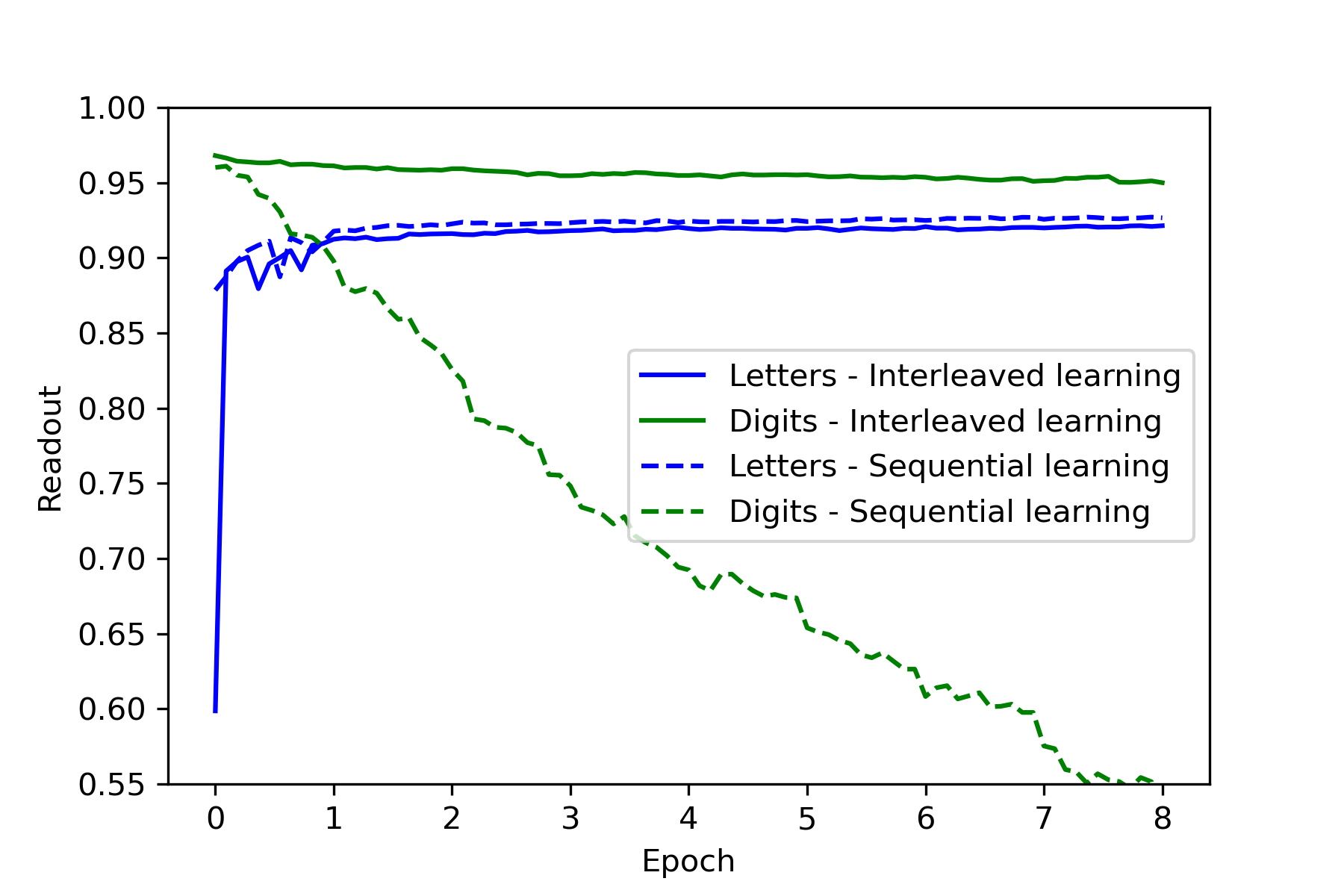}
    \caption{Readout accuracy in the continual learning scenario. The sequential learning regimen is strongly affected by catastrophic forgetting, while interleaved learning incorporates information from the new distribution (Letters) while preserving previous knowledge (Digits). Classifiers are evaluated at 10 regularly spaced intervals during each unsupervised learning epoch.}
    \label{fig:catastrphic}
\end{figure}

As clearly shown in Figure~\ref{fig:catastrphic}, the sequential learning setup (dashed lines) is dramatically affected by catastrophic interference: while the read-out accuracy on the new letter data set increases, the performance of the classifier trained on the previous digits data set steadily drops as learning proceeds. On the contrary, the interleaved learning setup (solid lines) allows to easily incorporate knowledge from the new letter data set, achieving the same accuracy of the sequential setup, at the same time allowing to maintain previously learned knowledge, as demonstrated by the preservation of the read-out accuracy for the digit data set.

\subsubsection{Emergence of structural properties during learning}
\begin{figure}
    \centering
    \begin{subfigure}[b]{0.95\textwidth}
        \includegraphics[width=\textwidth]{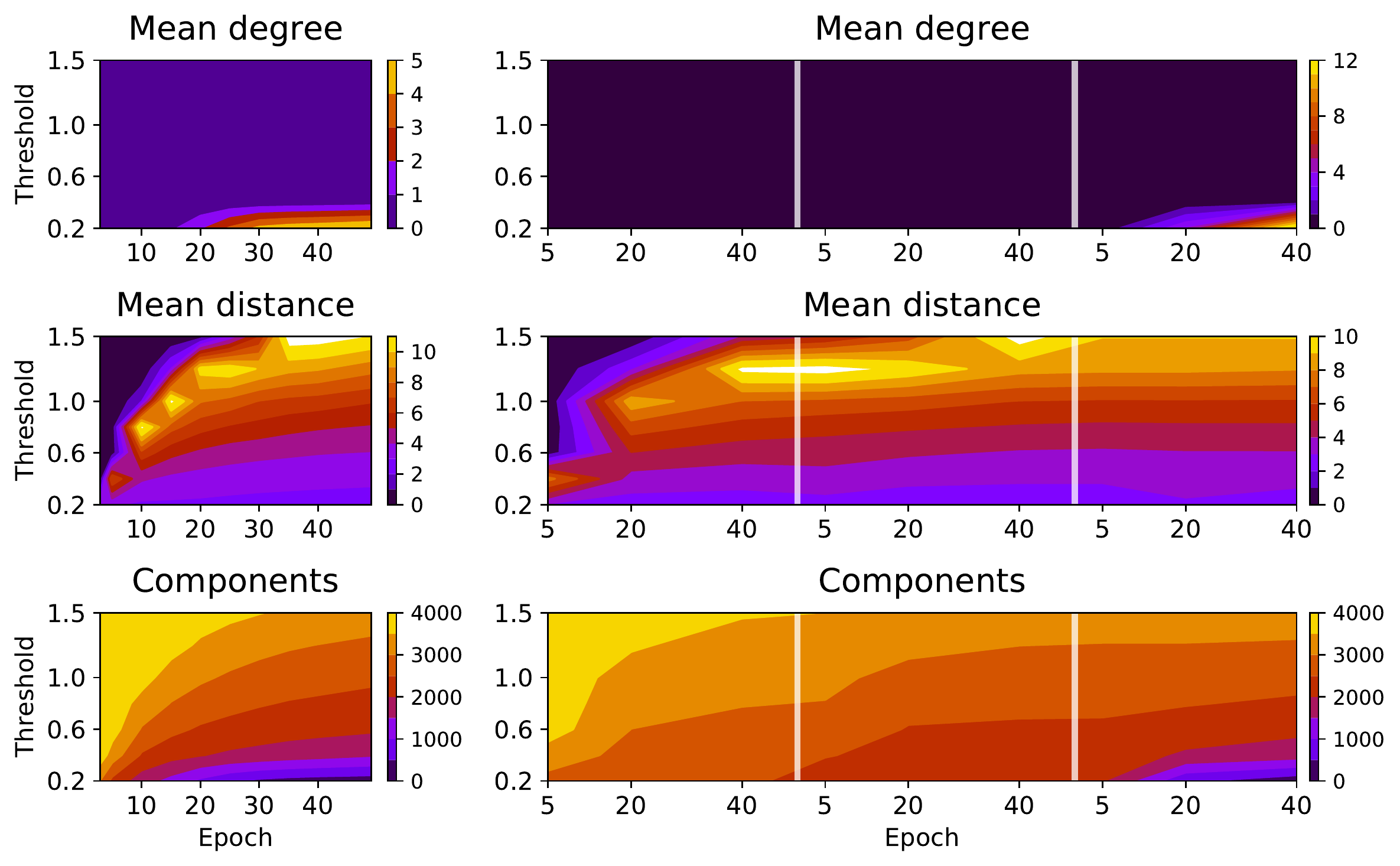}
        \caption{Real network. Left column: Iterative DBN; Right column: Greedy DBN.}
        \label{fig:kd_real_normal_nodrop}
    \end{subfigure}
    
    \begin{subfigure}[b]{0.95\textwidth}
        \includegraphics[width=\textwidth]{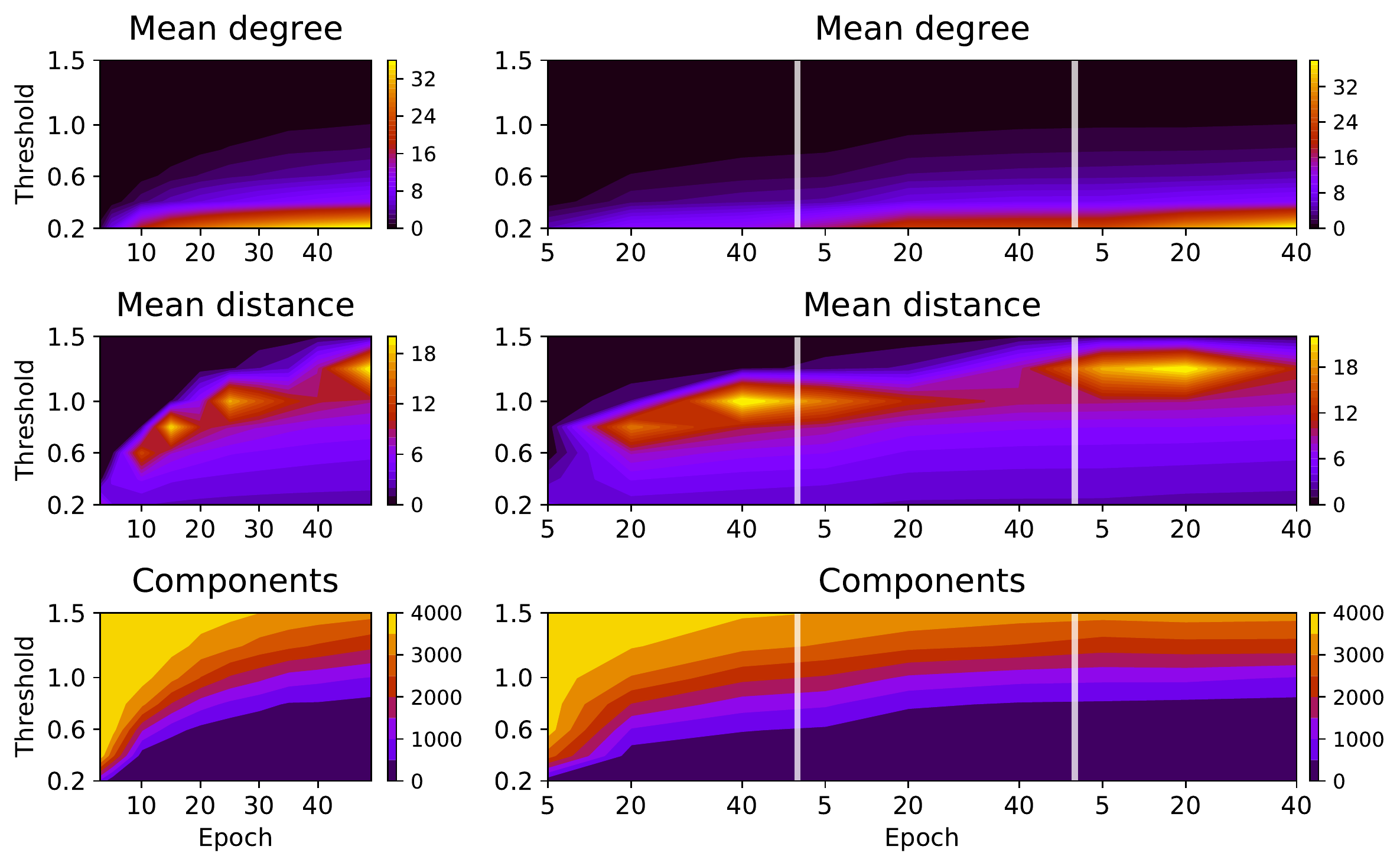}
        \caption{Binomial replica. Left column: Iterative DBN; Right column: Greedy DBN.}
        \label{fig:kd_fake_normal_nodrop}
    \end{subfigure}
    
    \caption{Contour plots of mean degree, mean geodesic distance and number of connected components for the case of Normal initialization without dropout. Note that while the iterative iDBN algorithm allows to analyze the entire network since earliest learning stages, in the greedy case the upper layers remain untouched by the update rule until the lower layers are fully trained. This motivates the visualization choice of the right column: the dashed line represents the subdivision between layers, so that to display the trend of change in the whole network during the true learning time-span.}
    \label{fig:kd_maps}
\end{figure}

Typically, inspecting the degrees distribution of a given network gives a first idea of the nature of the system, at least to determine whether the network is random or real. Here, our main focus is on how the structural properties (among the other, also the mean degree) are affected by the learning dynamics. The degree distribution itself is not informative about the structural evolution that the network experiences during training but still could give useful insights about the structural differences of each network. We chose to track other global properties, e.g. mean degree, mean geodesic distance and number of connected components. Figure~\ref{fig:kd_maps} displays these quantities, while Appendix~\ref{app:graphanalysis} provides a broad explanation on the degrees distribution and how to approach its evaluation in this constrained-architecture setup.

Results suggest that the deep network undergoes a substantial transformation during unsupervised learning. A major difference lays in the mean degree evolution during training. Referring to Figure~\ref{fig:kd_real_normal_nodrop}, non-trivial network properties tend to emerge in later stages of learning, especially for the greedy network. This suggests that the iterative implementation favors the development of complex circuits within the network connections and eventually the emergence of larger components. This applies both to the mean degree and to the number of connected components (recall that an isolated node itself is considered a component). The visualization of the mean geodesic distance shows that the evolution of this property is similar in both the greedy and iterative cases, in particular we can observe a phase transition between an initial state in which all nodes are isolated (indeed an isolated node is considered a component in which the geodesic distance is zero) to the emergence of some components. Once the different component connect to each other and the connections strengthen, the mean geodesic distance decreases.

The second set of results, in Figure~\ref{fig:kd_fake_normal_nodrop}, displays the evolution of the same properties in the binomial replicas. The names ``binomial greedy'' and ``binomial iterative'' mean that the binomial counterparts are obtained using the probabilities of edge existence derived from the real greedy and iterative networks, in the epochs considered. The main difference is that both the greedy and iterative instances display the same overall behavior. Unsurprisingly, the visualizations of mean degree and connected components practically overlap. This expected results is given by the fact that there is no such thing as the effect of a learning dynamics that shapes the networks internal structure. In addition, the bottom panels (connected components) show that in random networks the formation of only one giant component is strongly encouraged. Note further that mean degree and mean geodesic distances attain larger values in binomial networks. This is indeed what one could expect: random networks do not have \emph{hubs}, i.e. nodes with a larger degree with respect to the vast majority of the other nodes, which are instead a characteristic of real scale-free networks. The absence of few super-connected nodes implies a larger mean degree. 

\begin{figure}[t]
	\centering
	\begin{subfigure}[b]{0.475\textwidth}
		\includegraphics[width=\textwidth]{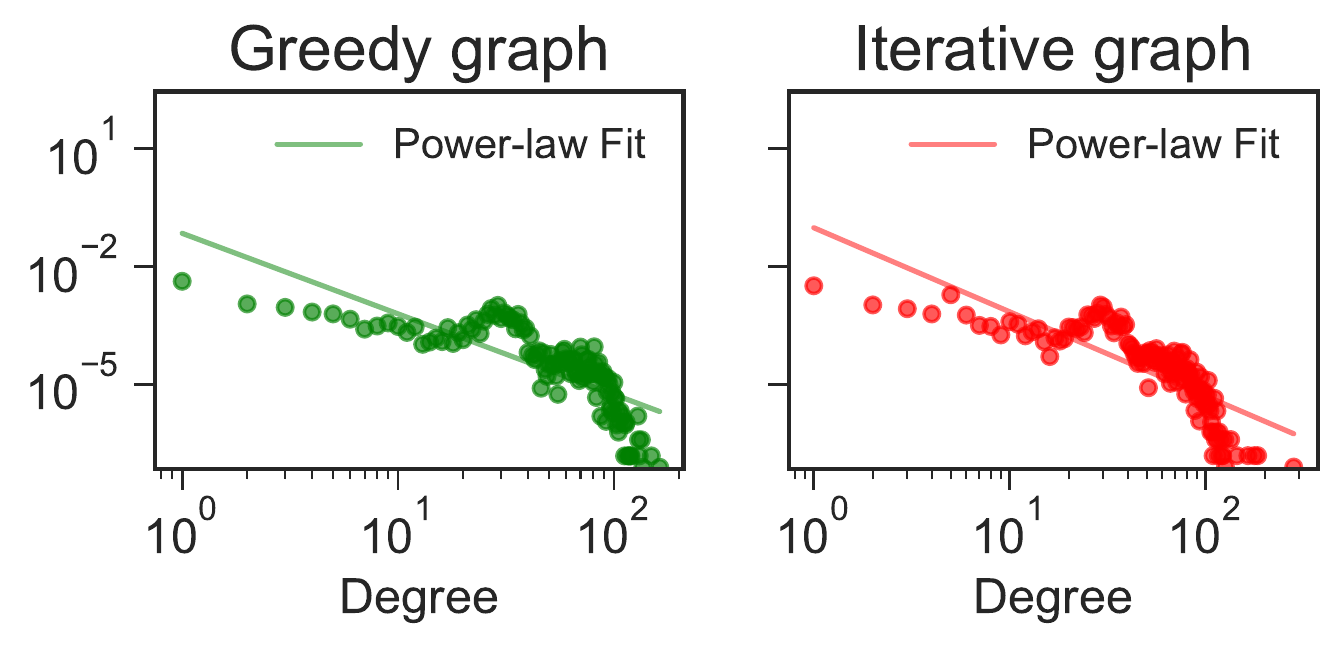}
		\caption{Real DBN, no dropout.}
		\label{fig:dds0.4_normal_real}
	\end{subfigure}
	~
	\begin{subfigure}[b]{0.475\textwidth}
		\includegraphics[width=\textwidth]{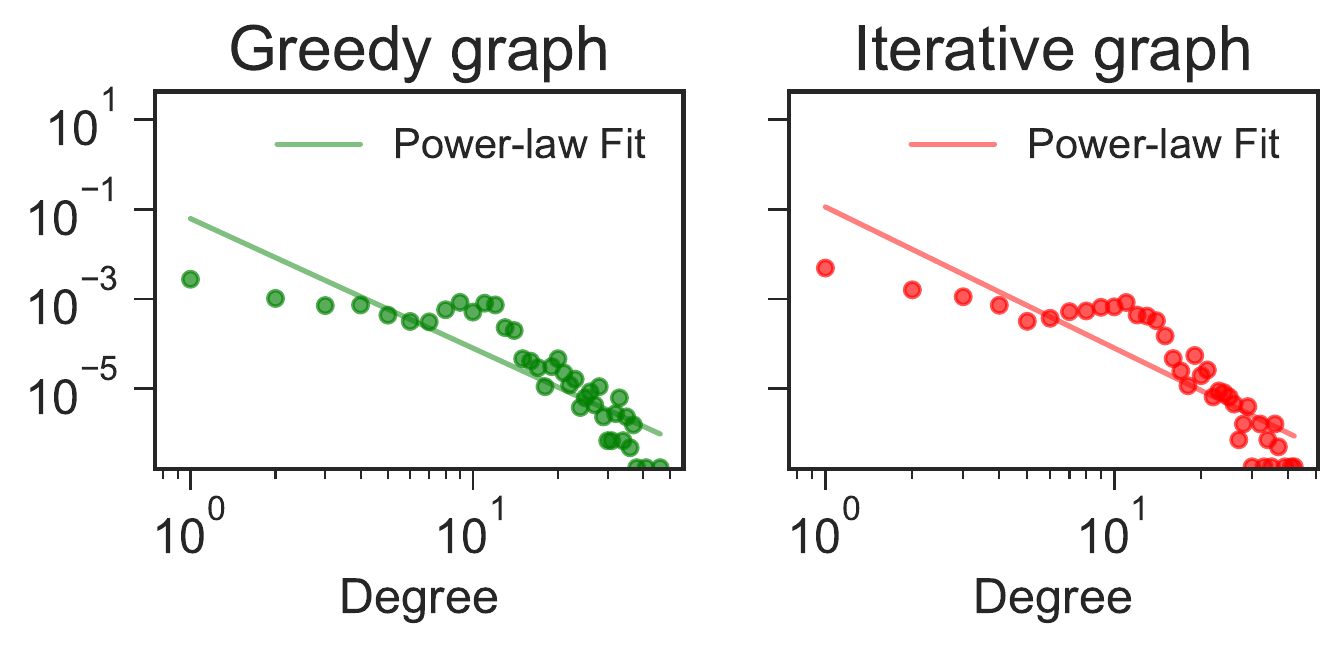}
		\caption{Real DBN, dropout with $p = 0.1$.}
		\label{fig:dds0.4_normal_real_0d1}
	\end{subfigure}
	
	\begin{subfigure}[b]{0.475\textwidth}
		\includegraphics[width=\textwidth]{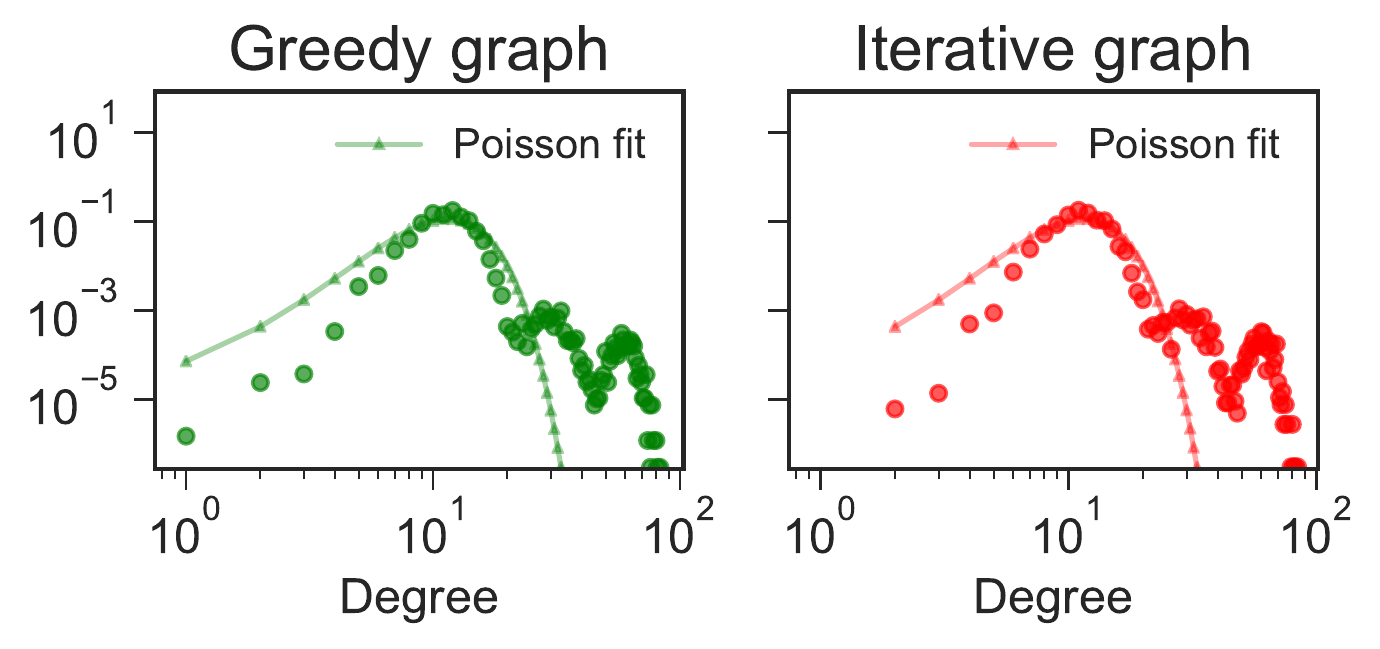}
		\caption{Binomial replica, no dropout.}
		\label{fig:dds0.4_normal_binomial}
	\end{subfigure}
	~
	\begin{subfigure}[b]{0.475\textwidth}
		\includegraphics[width=\textwidth]{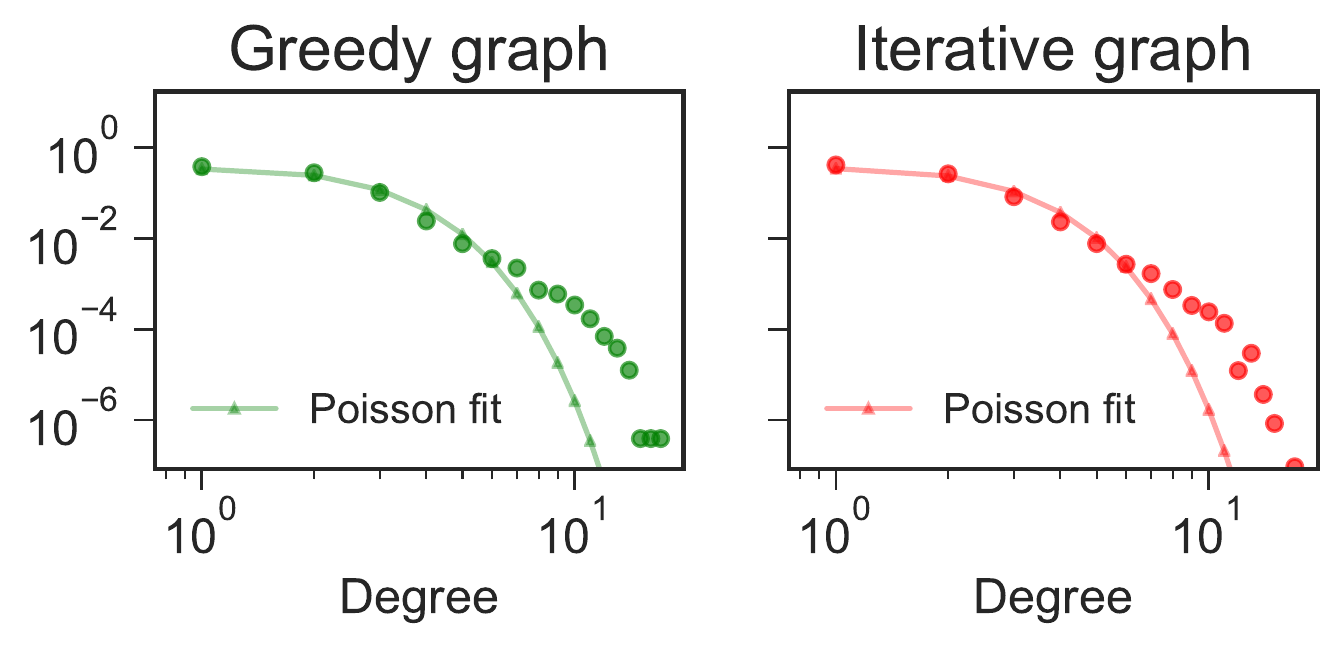}
		\caption{Binomial replica, dropout with $p = 0.1$.}
		\label{fig:dds0.4_normal_binomial_0d1}
	\end{subfigure}
	
	\caption{Degrees distributions with cut-off threshold set to $0.4$. The networks analysed for this figure have been initialized with the Normal distribution; the case of Glorot initialization does not display significant differences.}
	\label{fig:dds0.4}
\end{figure}

The results discussed above refer to the evolution of global network properties. As mentioned before, in network science it is well known that real networks exhibit a power-law degree distribution. As a further comparison between the real networks and their binomial counterparts, we thus choose to also inspect the degree distribution in both the greedy and iterative cases. For this test, we analyze the networks at the end of training and we also include the results obtained with the implementation of dropout, to see if the sparsity induced by dropout changes the network structure. Figure~\ref{fig:dds0.4} shows the visualization of the degree distributions for real and binomial graphs for the case of Normal initialization, both with and without dropout. The effect of the architectural bias discussed above is particularly clear on the two leftmost upper and lower panels, despite the implementation of the modified degree distribution that accounts for the maximal potential connectivity for each node. This result suggests that the modified degree distribution should be further refined in order to model more accurately the distribution of nodes degrees; still, the degrees distributions show qualitatively different shapes, suggesting that learning dynamics in DBNs characterizes the derived network as a non-random graph. Interestingly, the effect of dropout is to make random replicas more structurally similar to  real networks. The top and down right panels of Figure~\ref{fig:dds0.4} show that a larger connectivity for a small number of nodes is tolerated also in binomial graphs. 

\subsection{Numerosity data set}

\begin{figure}[t]
    \centering
    \begin{subfigure}[b]{0.45\textwidth}
        \includegraphics[width=\textwidth]{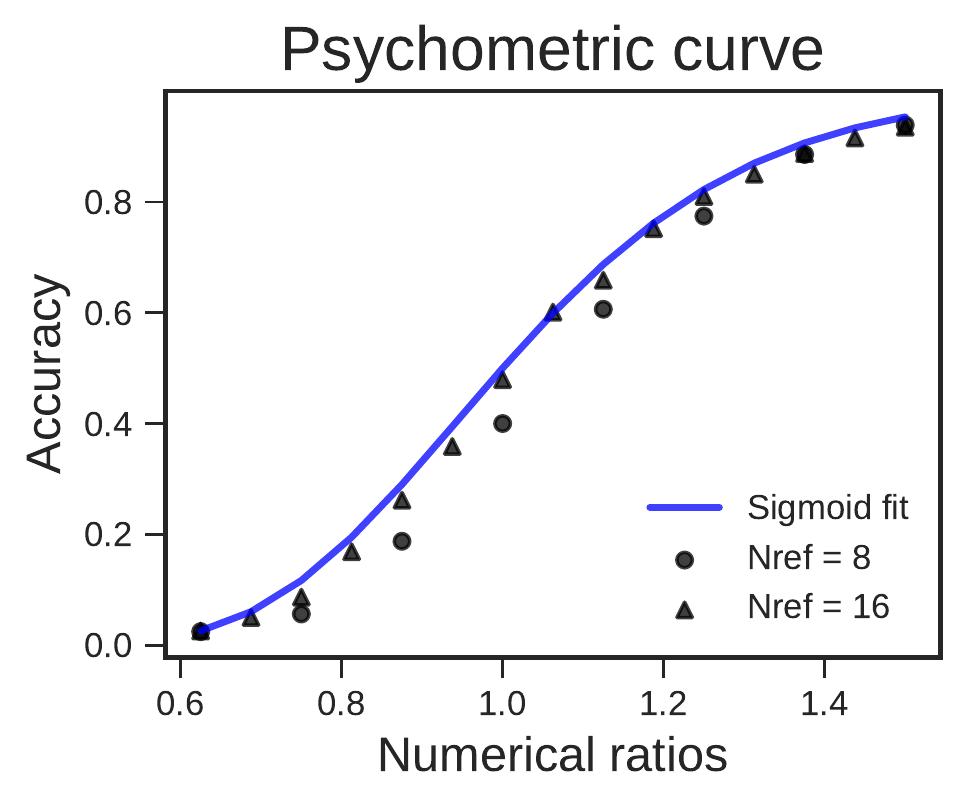}
        \caption{Sigmoidal fit of the psychometric function.}
        \label{fig:wsigm}
    \end{subfigure}
    ~
    \begin{subfigure}[b]{0.45\textwidth}
        \includegraphics[width=\textwidth]{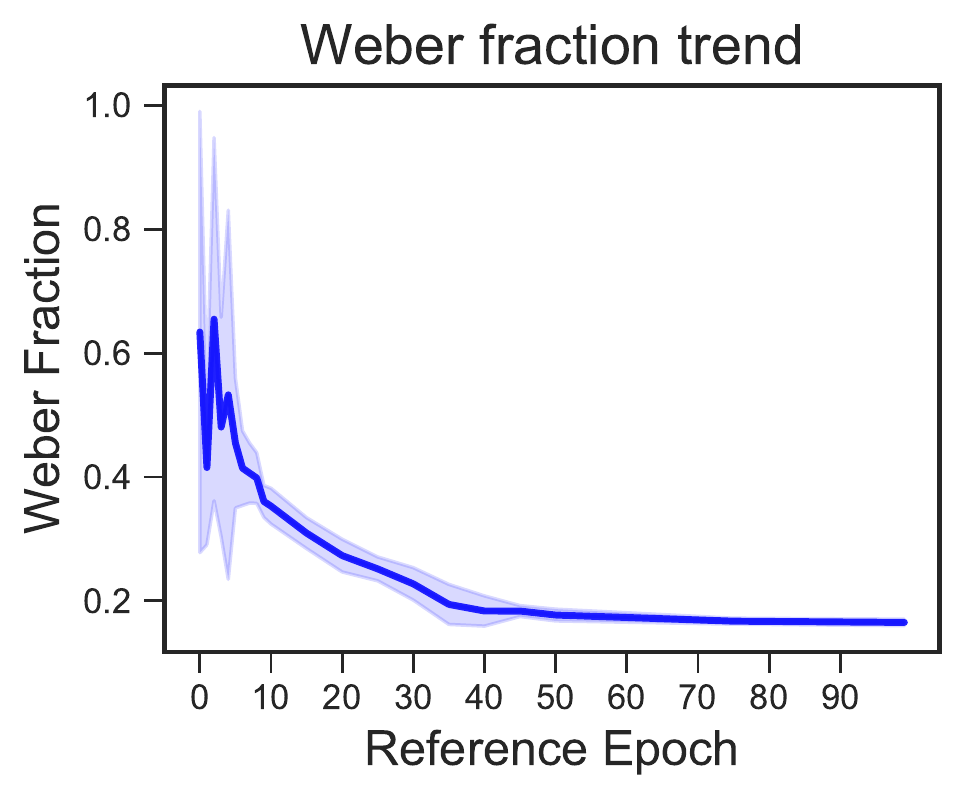}
        \caption{Trend of change of the Weber fraction.}
        \label{fig:wtrend}
    \end{subfigure}
    
    \begin{subfigure}[b]{0.45\textwidth}
        \includegraphics[width=\textwidth]{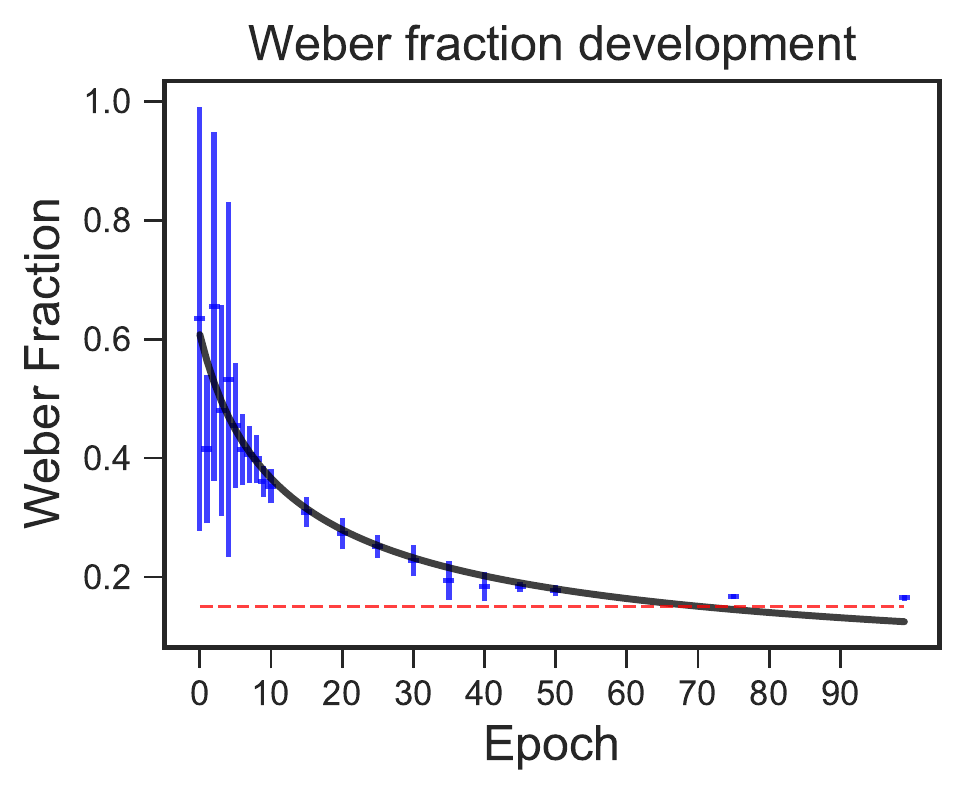}
        \caption{Values of $w$ as in Stoianov and Zorzi~\cite{stoianov2012}, fitted according to a power function.}
        \label{fig:wfit_sz}
    \end{subfigure}
    ~
    \begin{subfigure}[b]{0.45\textwidth}
        \includegraphics[width=\textwidth]{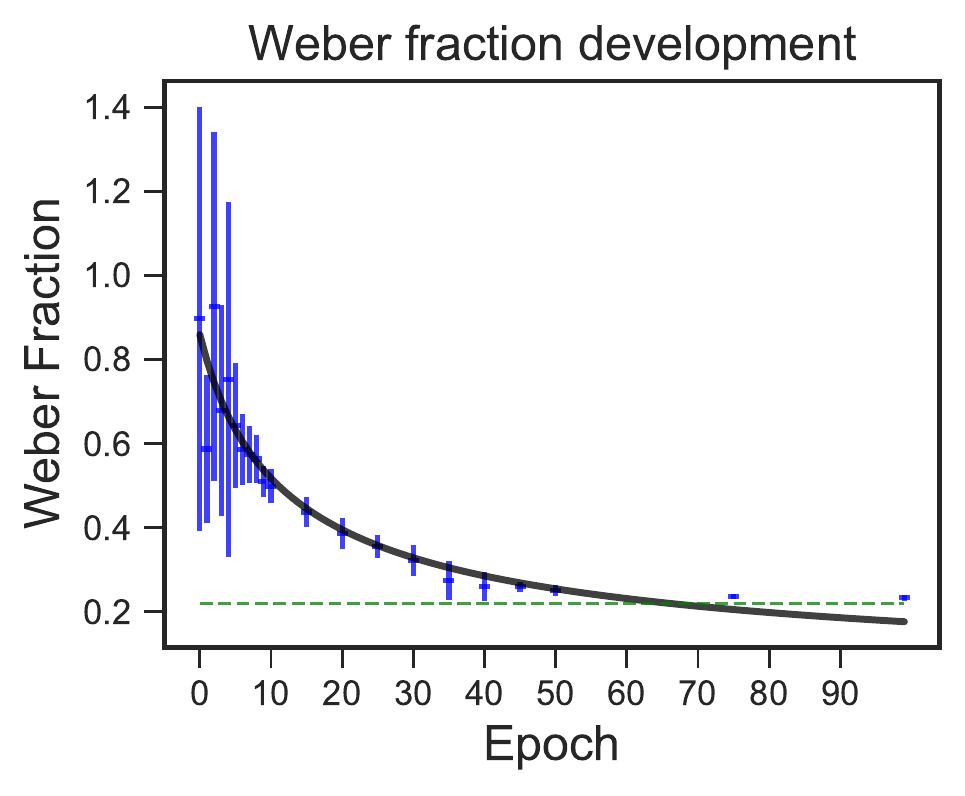}
        \caption{Values of $w$ rescaled as in Testolin et al.~\cite{testolin2020numerosity}, fitted according to a power function.}
        \label{fig:wfit_tzm}
    \end{subfigure}
    \caption{Final number acuity and developmental trajectories of the DBN. Panel~\ref{fig:wsigm} reports the psychometric curve obtained from the numerosity discrimination task at the end of unsupervised iDBN learning. The Weber fraction measures the steepness of the curve, here $w = 0.17$. Panel~\ref{fig:wtrend} reports the trend of decay of the Weber fraction during unsupervised learning. The solid curve is the average with respect to the runs of the average with respect to the classifier of the $w$ values, the shaded area is the standard deviation with respect to the runs. Panels~\ref{fig:wfit_sz} and~\ref{fig:wfit_tzm} show the power-law fit according to Equation \ref{eq:cabs}. The dashed lines represent reference values for the final $w$ from \cite{stoianov2012} and \cite{testolin2020numerosity} respectively.}
    \label{fig:devres}
\end{figure}

As shown in Figure~\ref{fig:wsigm} the psychometric function observed at the end of the unsupervised learning phase is well aligned with the results obtained by Stoianov and Zorzi using the greedy training scheme~\cite{stoianov2012}. The authors reported a Weber fraction value of $0.15$, while we obtained a value of $0.17$.

Concerning the developmental trajectory, as shown in Figure~\ref{fig:wtrend} the Weber fraction trend displays a significant decrease during the learning period, especially at the early stages of development. This trend is similar to that observed during human development \cite{halberda2008developmental,piazza2010developmental}. Note that our curve has been obtained by averaging $20$ model runs, training $5$ different classifiers for each data point to collect more reliable statistics. Thus, the average values account for a population of $100$ data points for each sample epoch. It is interesting to note that in the early stages the network might yield a worsening of the performance, and hence highly varying values of $w$. The behaviour stabilizes to an asymptotic value in more advanced learning stages.

Figures~\ref{fig:wfit_sz} and~\ref{fig:wfit_tzm} display the same data points, along with a power-law fit obtained using the method proposed by Testolin, Zou and McClelland~\cite{testolin2020numerosity}. Due to the initial zero value of the epoch time stamp, the basic functional form is actually a modified power-law \cite{reichert2013charles}:

\begin{equation}\label{eq:cabs}
    y = a(1 + s\,x)^{b}
\end{equation}

The parameters $a$, $b$ and $s$ are fitted to the data points describing the progressive development of the Weber fraction. The resulting parameters are reported in Table~\ref{tab:my_label}: the fit closely follows the trajectory of Weber fraction, hence describing satisfactorily well the development of the number sense across the learning period of our networks.

\begin{table}[!h]
    \centering
    \begin{tabular}{c c c c c}
        \hline
                & $a$    & $b$    & $s$    & $R^2$ \\
        \hline
         S$\&$Z & $0.61$ & $0.59$ & $0.13$ & $0.9$ \\
         TZM    & $0.86$ & $0.59$ & $0.13$ & $0.9$ \\
        \hline
    \end{tabular}
    \caption{Fitted parameters as in Equation~\ref{eq:cabs} for both the $w$ values scaled, respectively, according to the method of S$\&$Z \cite{stoianov2012} and TZM \cite{testolin2020numerosity}.}
    \label{tab:my_label}
\end{table}

\section{Discussion}
\label{sec:discussion}

Our simulations demonstrate that deep belief networks can be trained iteratively, by jointly adjusting all the weights of the model hierarchy following observation of each sensory pattern (or minibatch of patterns). This innovative learning algorithm can be used in place of the traditional greedy, layer-wise learning algorithm in order to accurately track the developmental trajectory of the model. This allows to study how global properties of the network can gradually emerge during the course of learning, at the same time enabling a systematic comparison with biological developmental trajectories observed in empirical studies.

The proposed algorithm was first evaluated on the popular MNIST benchmark, where DBNs trained using the greedy algorithm have traditionally achieved very good performance. We showed that the DBN trained using our iterative algorithm was able to achieve a performance comparable to the greedy counterpart, both in terms of readout accuracy from the internal representations and in terms of reconstruction capabilities. We also probed the final models on a variety of generative tasks, which assessed the DBN ability to reproduce, complete and denoise corrupted input images. Also in this case, we did not observe significant differences between the greedy and iterative versions of the learning algorithm. On the contrary, the attempt to implement joint training of all weights by simply propagating signals across the full stack both in the bottom-up and top-down learning phases led to poor convergence and unsatisfactory results. This reveals that, subsequently to the feed-forward propagation across the entire DBN hierarchy (which resembles the fast feed-forward sweep observed in cortical circuits \cite{lamme2000distinct,vanrullen2007power}), neurons' activation at the deepest layer cannot be fed all the way back to the visible layer in a (symmetrical) fast feedback sweep but need to be locally processed at each layer to compute the learning signals (as implemented in the iDBN). Notably, the latter scheme is also consistent with local recurrent processing supported by recurrent and horizontal connections within cortical areas \cite{lamme2000distinct,kreiman2020beyond}.

To further support the cognitive plausibility of the proposed developmental scheme, we also implemented a straightforward interleaved training approach to tackle continual learning tasks, demonstrating that the iDBN can be also extended to challenging scenarios that require to incrementally incorporate new knowledge in the deep network. \remembertext{Comment-1.4.4}{This paves the way to the investigation of more plausible continual learning schemes that exploit the top-down generative properties of the DBN to sample the stimuli that are needed for interleaved training \cite{calandra2012}.} Moreover, we carried out an extensive analysis on the progressive development of structural properties in the DBN, by investigating the emergence of a variety of graph-theoretical properties, such as degree, geodesic distance and connected components. Our iterative learning approach allowed to emphasize the gradual refinement of these properties at the global level, thus opening the possibility to more systematically study the topological development of such complex hierarchical systems.

We finally evaluated our iterative approach on a  data set that has been recently exploited in a variety of cognitive models to simulate the perception of visual numerosity. Also in this case, the iDBN achieved a final accuracy comparable to the greedy counterpart, at the same time allowing for a precise tracking of the progressive development of the internal representations of the model. Behavioral performance, supported by a linear read-out from the internal representations, can also be continuously tracked to monitor skill acquisition. Indeed, the learning curves resulting from our approach closely resemble the learning curves reported in experimental studies with human children, and also overlap with those reported in a recent developmental model based on deep autoencoders.

\section{Conclusion}
\label{sec:conclusion}
The scope of the present work was twofold. On the one hand, we presented a novel unsupervised learning algorithm for deep belief networks that allows to accurately track the progressive development of the internal representations of the model. We validated our algorithm on two prototypical benchmark domains, achieving results comparable to the state-of-the-art. On the other hand, we demonstrated that our iterative learning algorithm can be extended to more realistic learning scenarios, at the same time supporting the psychometric analysis of progressive changes in behavioral performance and the study of the gradual development of network properties from the point of view of network theory.

Future studies might take advantage of the proposed iterative algorithm to investigate how the emergence of developmental disorders can be related to impairments in both the initial conditions of the system and the subsequent learning phases. This would be particularly relevant for the study of widespread learning disabilities such as dyscalculia, for which we are still lacking a computational characterization \cite{zorzitestolin2022}.
It would also be useful to explore whether the proposed iDBN approach could be applied in other cognitive domains involving unsupervised deep learning. This would be valuable not only from a machine learning standpoint, but also from a cognitive modeling perspective: Indeed, our iterative algorithm allows to create developmental models that can be quantitatively validated against empirical data collected on humans. Similarly, the proposed graph analysis could be applied in developmental physiology to better understand how structural and functional properties of self-organizing networks might gradually emerge from unsupervised learning dynamics.

\newpage
\begin{appendices}
\counterwithin{figure}{section}

\section{Supplementary Figures}

\begin{figure}[h]
    \centering
    \includegraphics[width=0.95\textwidth]{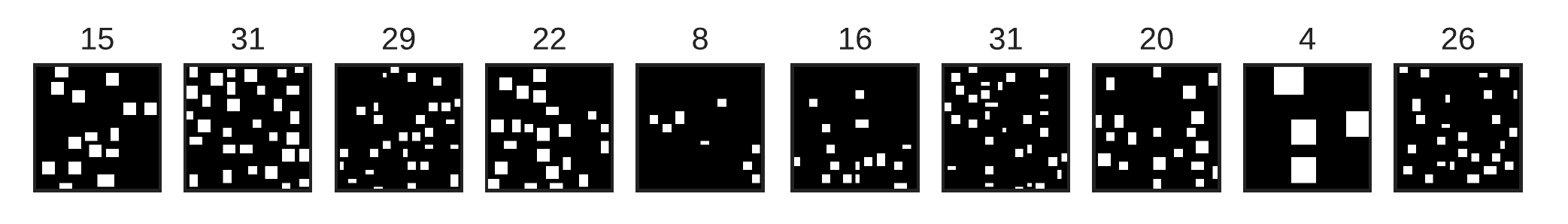}
    \caption{Samples from the Numerosity data set, which contains 51200 images featuring a variable number of white rectangles drawn on a black background. Numerosity ranges from $1$ to $32$ and objects have variable position and dimension (see~\cite{stoianov2012} for further details). The corresponding numerosity is reported on top of each image.}
    \label{fig:sz}
\end{figure}

\begin{figure}[h]
    \centering
    \includegraphics[width=0.7\textwidth]{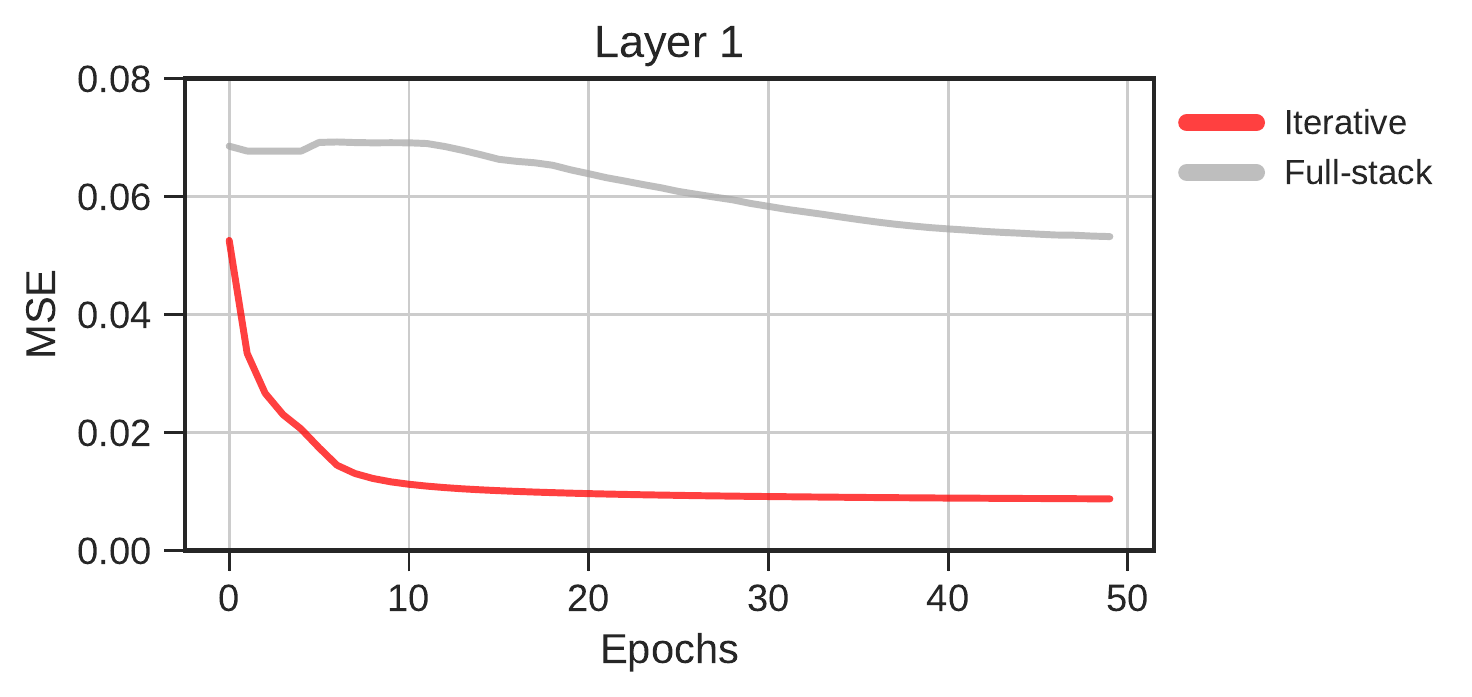}
    \caption{Reconstruction error trend of the first hidden layer for the iterative (iDBN) vs. full-stack developmental schemes. We verified that convergence for the full-stack scheme did not improve even after prolonging learning for 100 epochs.}
    \label{fig:iterative_vs_fullstack}
\end{figure}

\begin{figure}[]
    \centering
    
    \begin{subfigure}[b]{0.45\textwidth}
        \includegraphics[width=\textwidth]{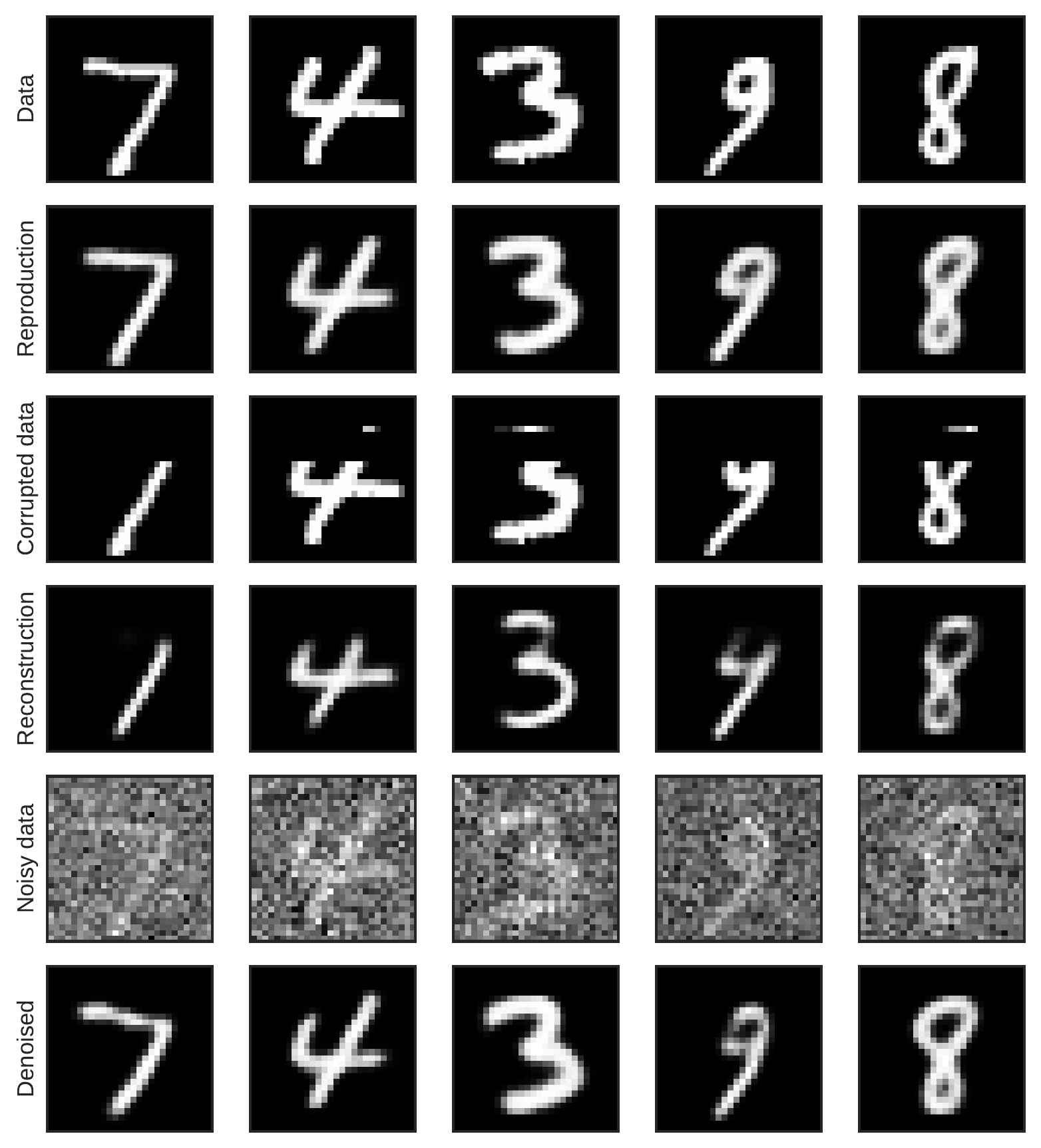}
        \caption{Greedy DBN.}
    \end{subfigure}
    ~
    \begin{subfigure}[b]{0.45\textwidth}
        \includegraphics[width=\textwidth]{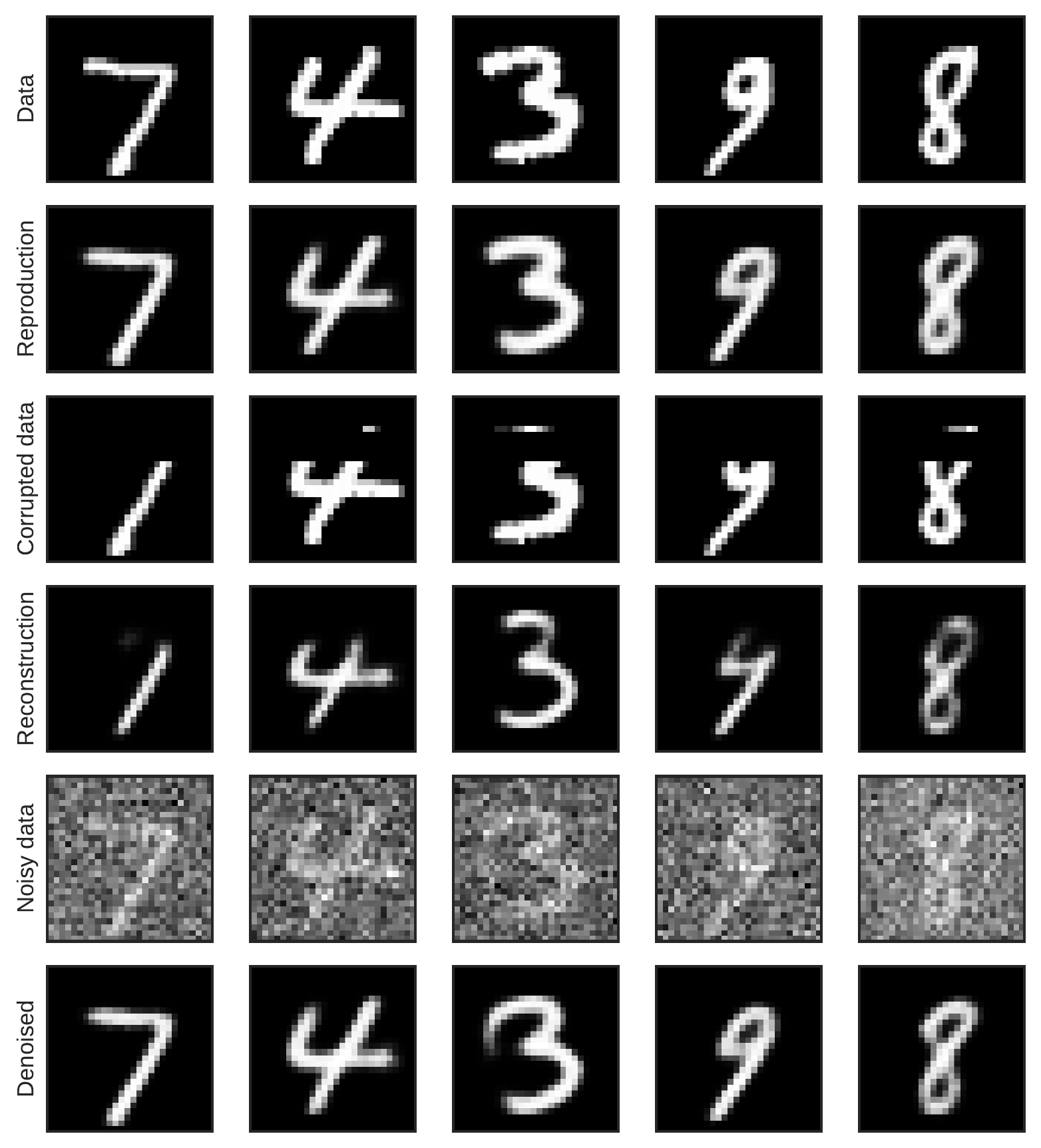}
        \caption{Iterative DBN.}
    \end{subfigure}
    
    \caption{Examples from the image generation tasks. First row: original data. Second row: reproduction of the original data. Third row: Partially observed data. Fourth row: completion of partially observed data. Fifth row: noisy data. Sixth row: denoised data.}
    \label{fig:data-reconstruction}
\end{figure}

\begin{figure}[]
    \centering
    \begin{subfigure}[b]{0.25\textwidth}
        \includegraphics[width=\textwidth]{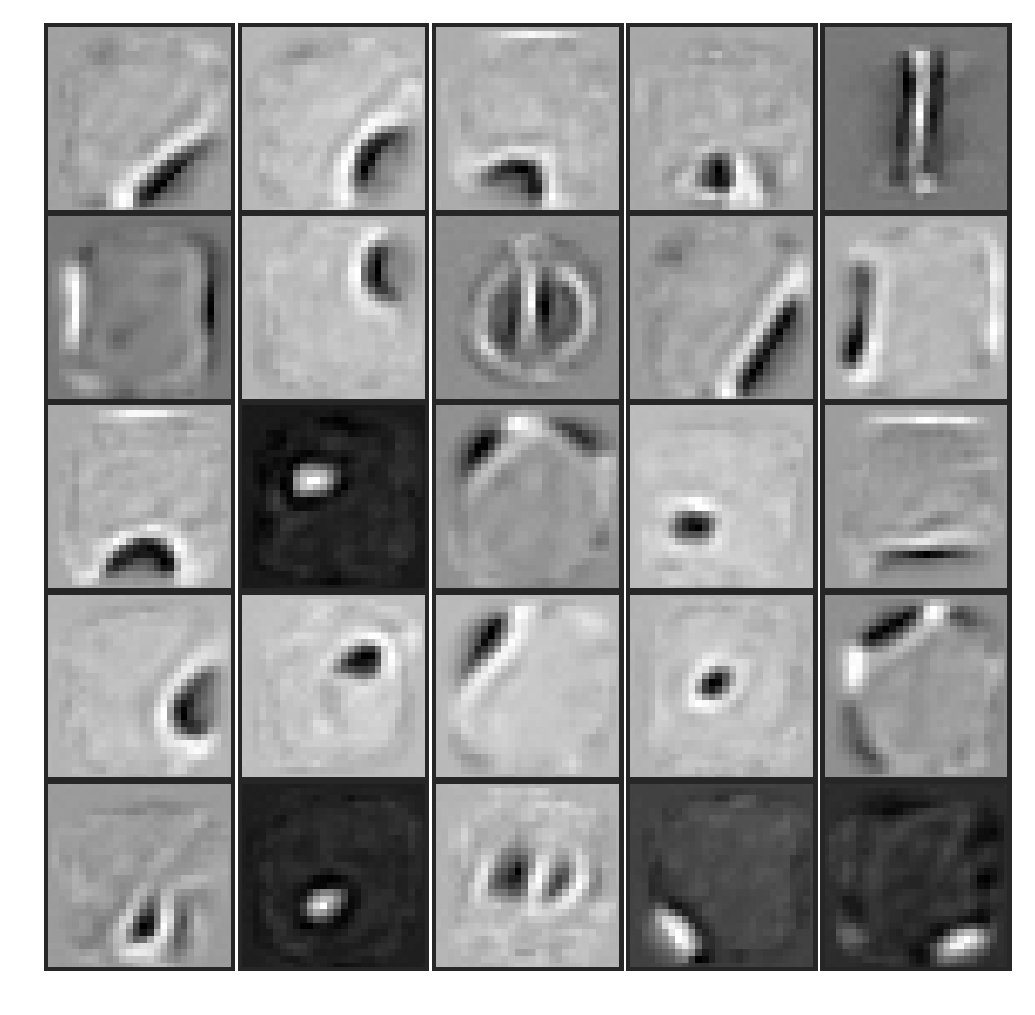}
        \caption{Layer 1.}
    \end{subfigure}
    ~
    \begin{subfigure}[b]{0.25\textwidth}
        \includegraphics[width=\textwidth]{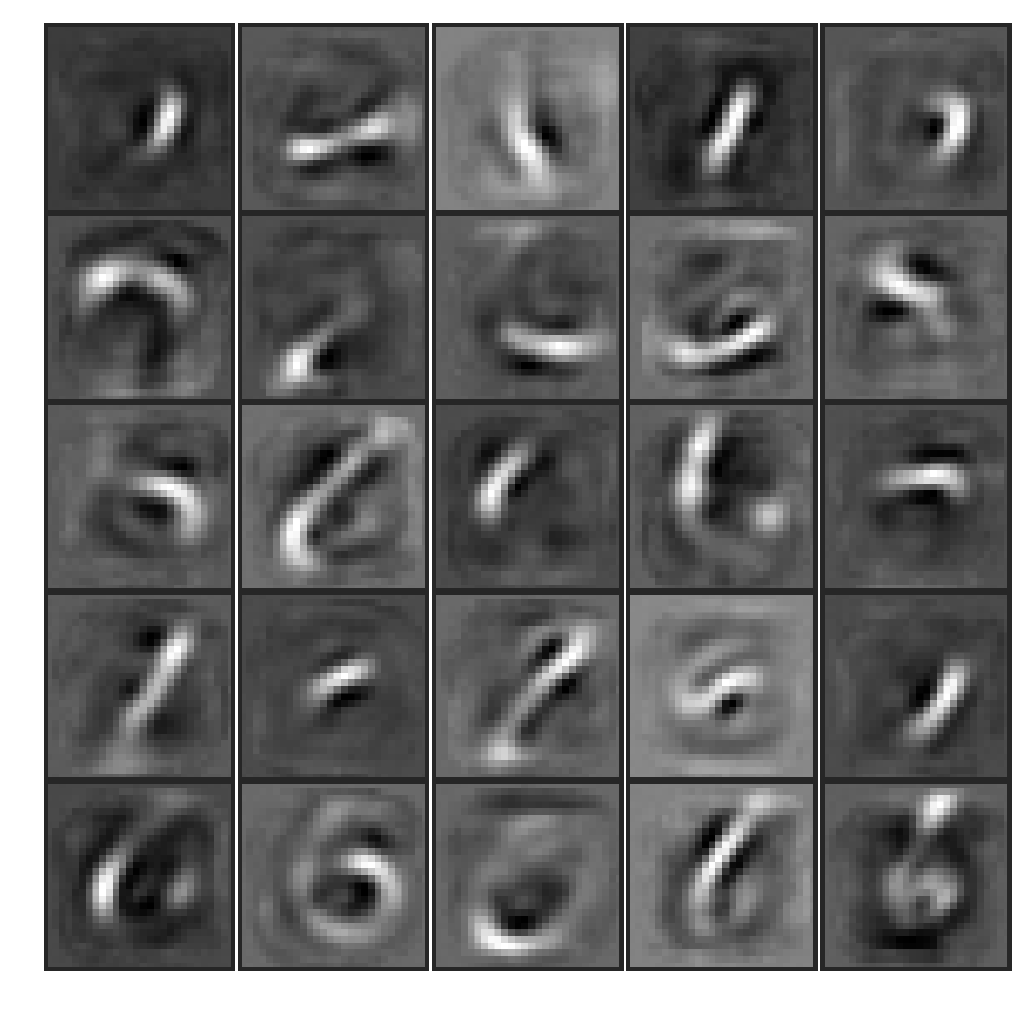}
        \caption{Layer 2.}
    \end{subfigure}
    ~
    \begin{subfigure}[b]{0.25\textwidth}
        \includegraphics[width=\textwidth]{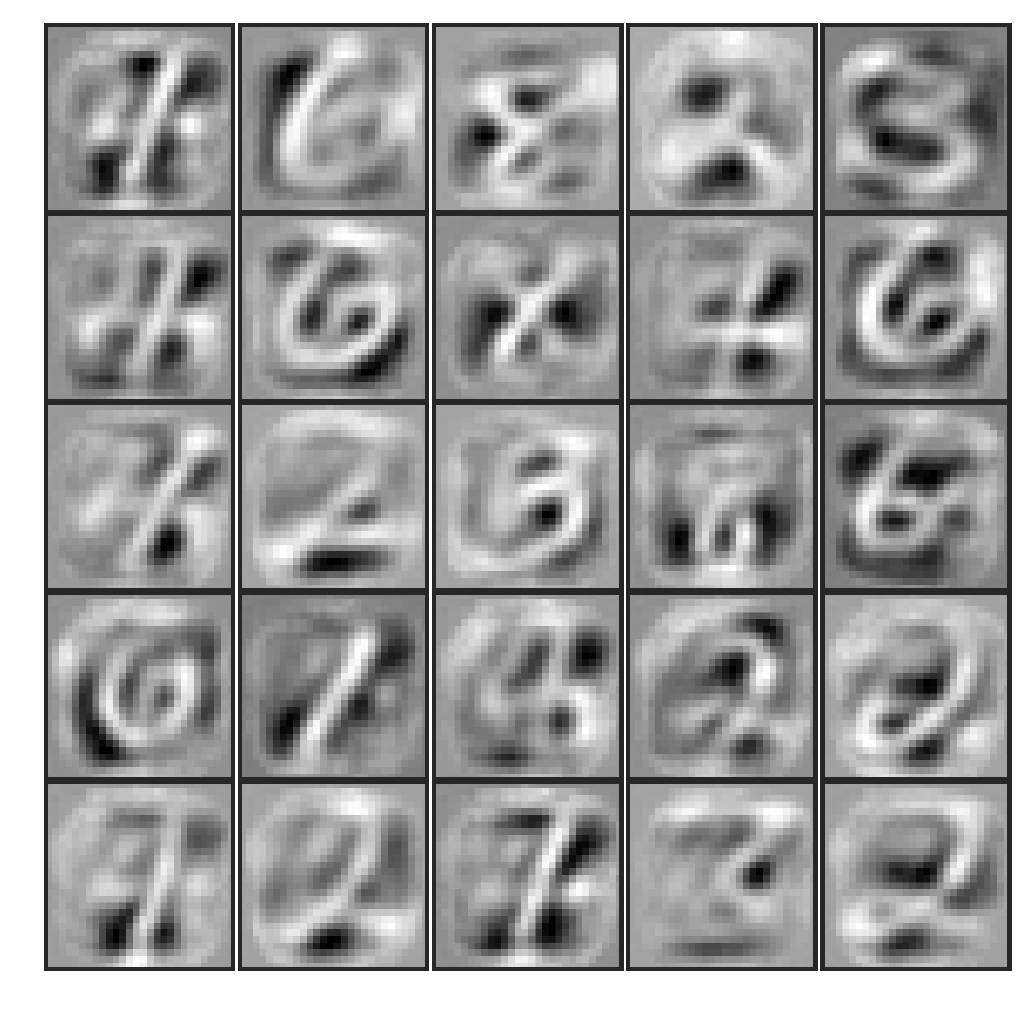}
        \caption{Layer 3.}
    \end{subfigure}
    
    \begin{subfigure}[b]{0.25\textwidth}
        \includegraphics[width=\textwidth]{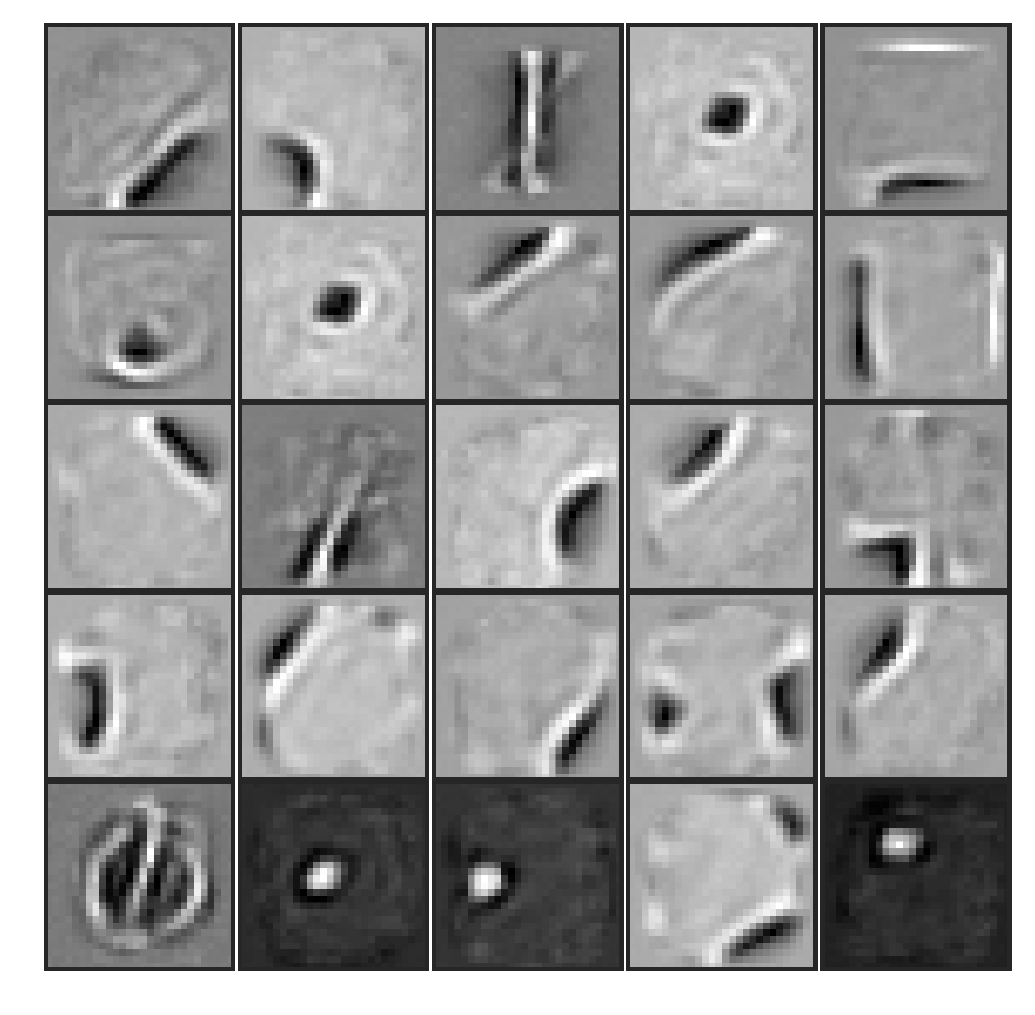}
        \caption{Layer 1.}
    \end{subfigure}
    ~
    \begin{subfigure}[b]{0.25\textwidth}
        \includegraphics[width=\textwidth]{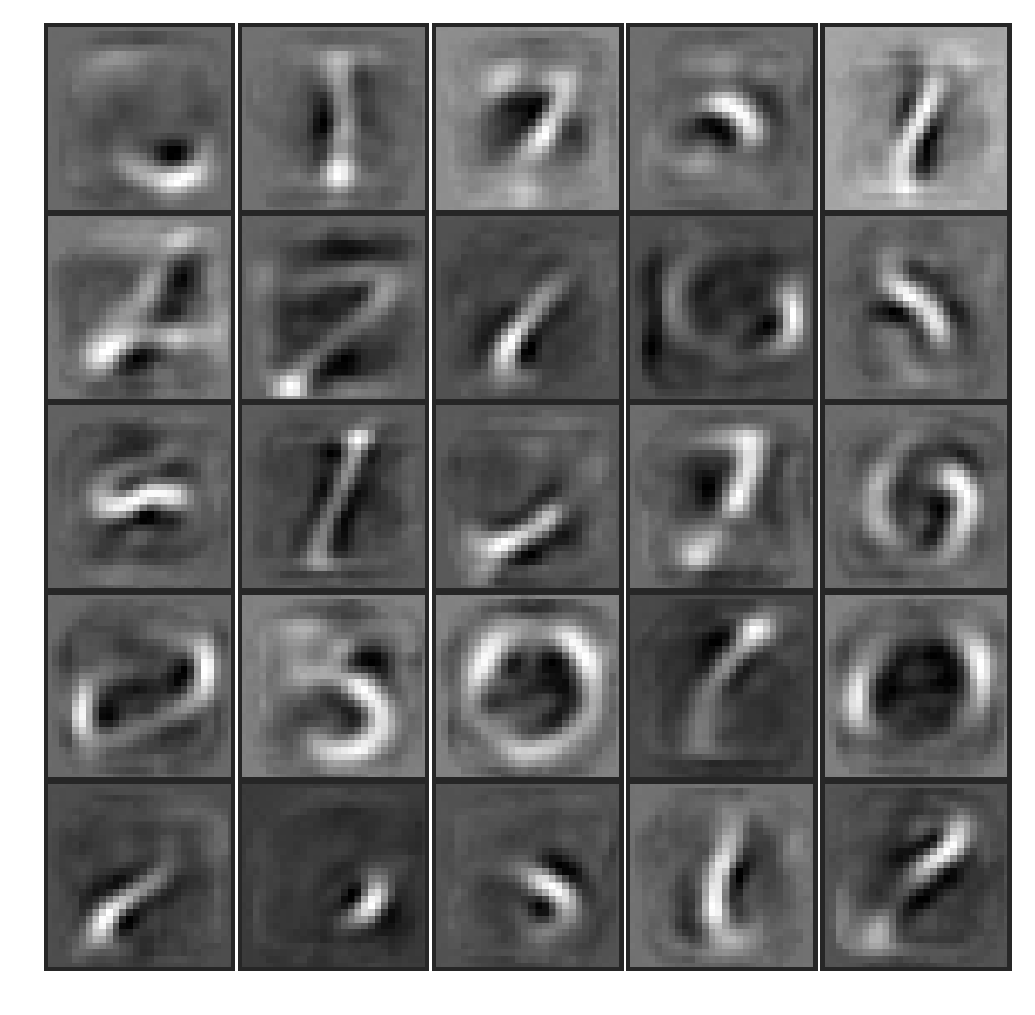}
        \caption{Layer 2.}
    \end{subfigure}
    ~
    \begin{subfigure}[b]{0.25\textwidth}
        \includegraphics[width=\textwidth]{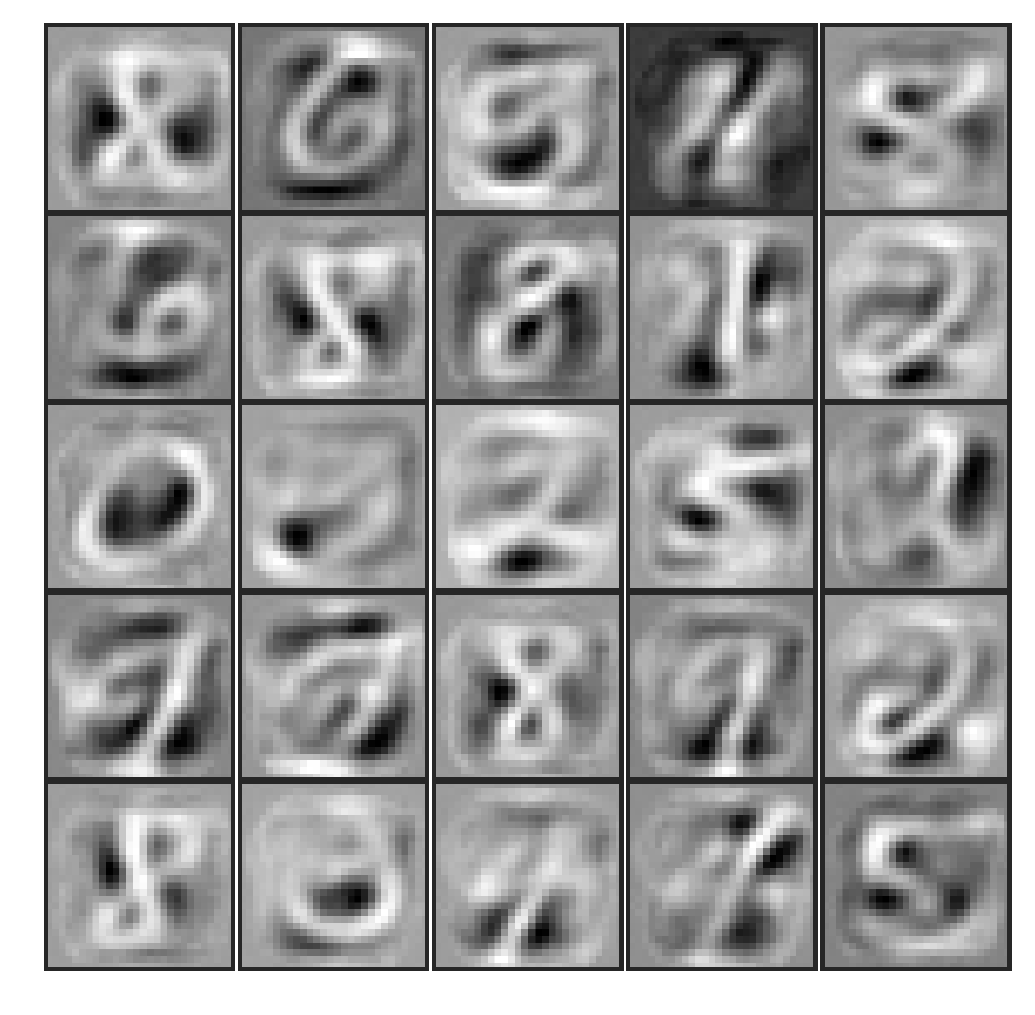}
        \caption{Layer 3.}
    \end{subfigure}
    
    \caption{Emerging receptive fields for the Normal initialization. Top row: Greedy algorithm. Bottom row: Iterative algorithm.}
    \label{fig:receptivefields_normal}
\end{figure}

\newpage
\section{Robustness of the iterative learning scheme results}
\label{app:robustness}
Results discussed in the main text about the equivalence in terms of performance between greedy and iterative training were focused on the Normal initialization configuration, with no dropout. To assess the robustness of our analyses, here we show that the same results are found when the weights are sampled à la Glorot and when dropout is added as a regularizer.

\subsection{Glorot initialization and dropout}
\begin{figure}[h]
    \centering
    \captionsetup{justification=centering}
    
    \begin{subfigure}[b]{0.9\textwidth}
        \includegraphics[width=\textwidth]{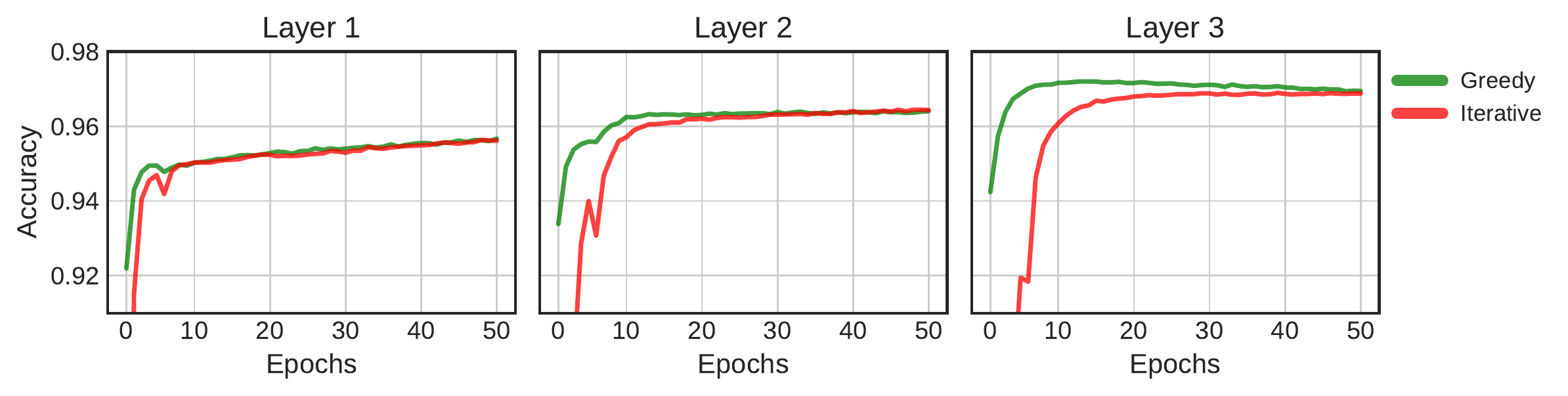}
        \caption{Readout profiles at different layers of the DBN hierarchy during training.}
        \label{fig:cd1_readout_glorot}
    \end{subfigure}
    
    \begin{subfigure}[b]{0.9\textwidth}
        \includegraphics[width=\textwidth]{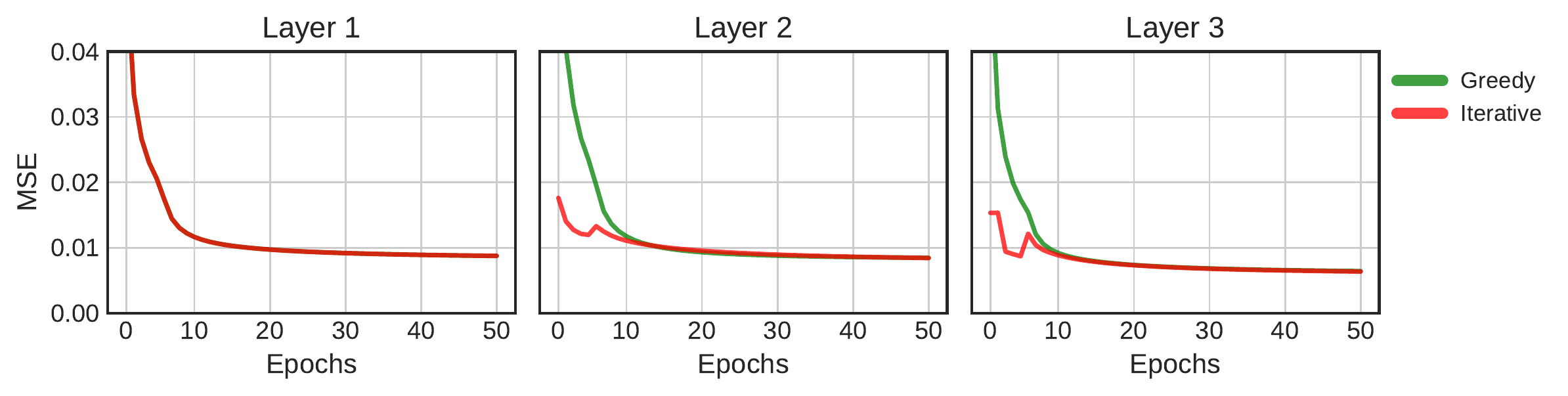}
        \caption{Reconstruction errors at different layers of the DBN hierarchy during training.}
        \label{fig:cd1_cost_glorot}
    \end{subfigure}
    
    \begin{subfigure}[b]{0.31\textwidth}
        \includegraphics[width=\textwidth]{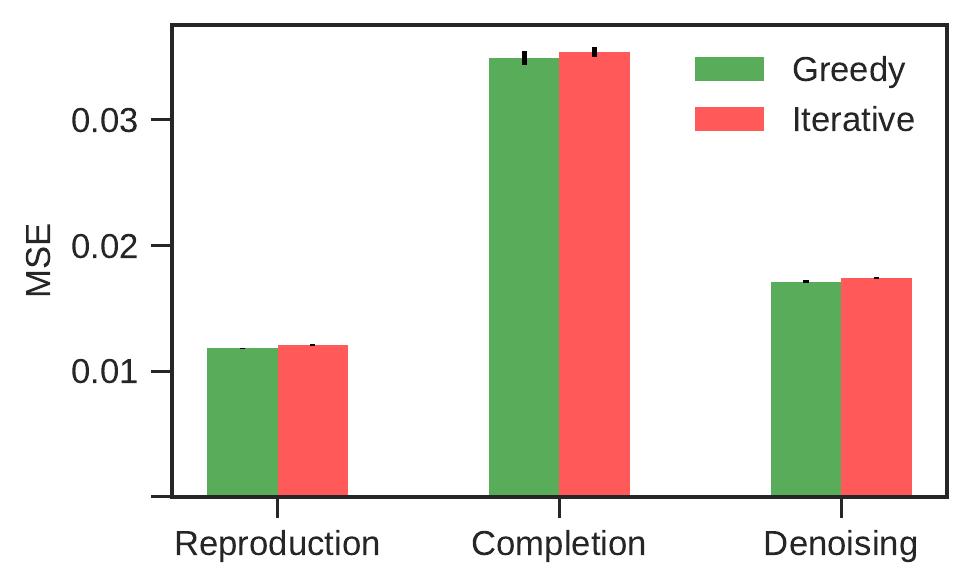}
        \caption{Generation tasks\\ (Glorot, no dropout)}
        \label{fig:cd1_mses_glorot}
    \end{subfigure}
    \begin{subfigure}[b]{0.31\textwidth}
		\includegraphics[width=\textwidth]{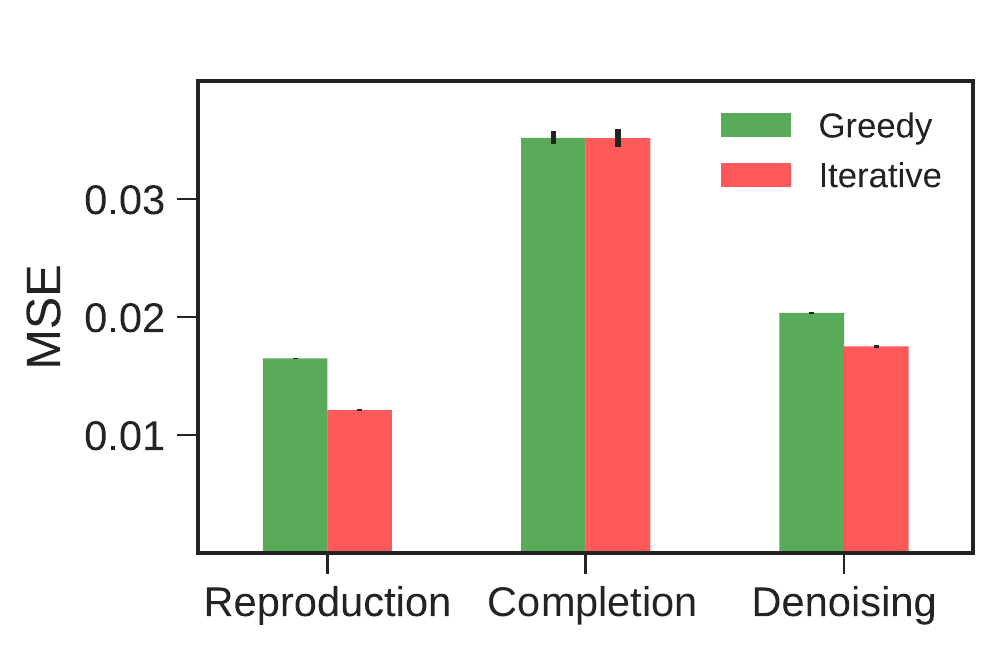}
		\caption{Generation tasks\\ (Normal, dropout)}
	\end{subfigure}
	\begin{subfigure}[b]{0.31\textwidth}
		\includegraphics[width=\textwidth]{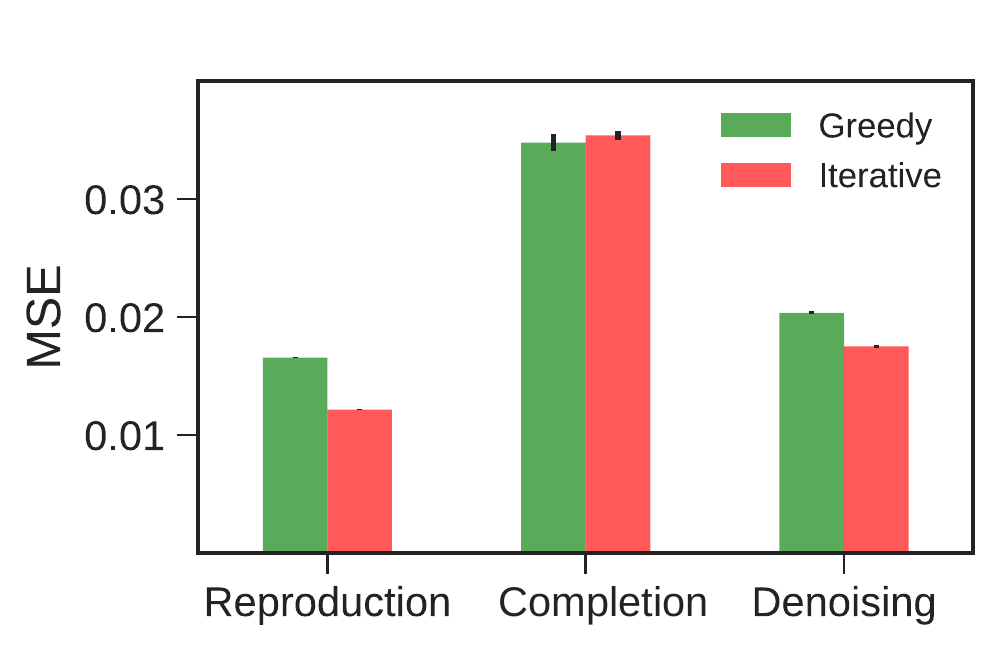}
		\caption{Generation tasks\\ (Glorot, dropout)}
	\end{subfigure}
	
    \caption{(a-b) Performance of the greedy vs. iterative schemes during learning, for the Glorot weights initialization. (c-e) Generative accuracy of the greedy vs. iterative schemes at the end of learning, for combinations of Glorot initialization and dropout.}
    \label{fig:learning-curves-glorot}
\end{figure}

Figure~\ref{fig:learning-curves-glorot} reports the readout and reconstruction profiles obtained with the Glorot initialization. The results show consistency with those presented in the main text. As mentioned before, the weight matrices are down-scaled by a factor $0.1$ in such a way to bring these weights values in a range similar to the weights of Normal initialization. 

We also evaluated the effect of dropout on our training scheme. Dropout is an effective regularization method that can be applied to RBMs \cite{srivastava2012}. The idea is to randomly silence some neurons during training, in such a way to prevent connections to over-learn. This can yield improved generalization capability, though in some cases it might also hinder learning performance. Bottom panels in Figure~\ref{fig:learning-curves-glorot} shows that also in this setup the overall performance of the model is not affected.   

\newpage
\section{Graph analysis in depth}\label{app:graphanalysis}
In Section~\ref{sec:resultsgraph} we discussed the critical choice of the cut-off threshold and displayed aggregated data of the emerging global structural properties. Here the degrees distributions are discussed in finer-grained detail, since the graph structure extrapolated from a DBN poses some case-specific problems. Indeed, it is not straightforward to choose a model distribution to fit the emergent weights configuration, since during learning some connection strengths are significantly increased, leading to long-tail frequency distributions. Rather than searching or producing an ad-hoc probability distribution for the learned weights and then pruning the network according to some distribution-specific values (e.g. quantiles), an heuristic choice is rather to prune the network based on a user-defined cut-off threshold. In Section~\ref{sec:resultsgraph}, a set of such meta-parameters is used jointly with the reference epochs to draw mean degrees, distances and connected components maps. Here the focus is on the degree distribution for a given cut-off threshold, which has different characteristics in random graphs and real networks~\cite{barabasi2016}. 

\subsection{Degrees distribution}
Real networks typically exhibit a degree distribution than follows a power-law, expressed by:

\begin{equation}
	p_k = a \, k^{-\gamma}
\end{equation}
where $k$ represents the degree, $\gamma$ is the decaying exponent and $a$ is a constant. Results in Figure~\ref{fig:dds0.4} were obtained by eliminating all the weights in the interval $\left[-0.4, 0.4\right]$, thus excluding the majority of connections. The power-law fit performs poorly for these DBN-like networks and the shape of the distribution deviates from the linear trend typical of scale-free networks, creating issues with the empirical way of computing the degrees distribution. Typically, one would count all the nodes having degree $k$, call this quantity $N_k$. The fraction of such nodes is:

\begin{equation}
    p_k = \dfrac{N_k}{N}
\end{equation}
where $N$ is the number of nodes in the network and $N_k$ is the fraction of nodes having degree $k$. This expression identifies a legitimate probability mass function, since $\sum_k p_k = 1$. The network architecture poses a subtle problem: while in real networks any node could be connected to virtually any other, here each node can be connected to all and solely those belonging to neighboring layers. Thus, each node should be associated with a potential maximum degree. This quantity is used to correct the degrees distribution, by penalizing a node depending on its maximum allowed connectivity, thus smoothing the resulting distribution. The nodes fraction is set to:

\begin{equation}
\label{eq:ddcorr}
    \Tilde{p}_k = \dfrac{N_k}{N} \sum_{i = 1}^{N} \mathbb{I}_{k_i = k} \frac{1}{\sum_{j = 1}^{N} A_{ij}}
\end{equation}
where $\mathbb{I}$ is the indicator function. This distribution is normalized computing the normalizing function numerically as $C = \sum_k \tilde{p}_k$. Assume that node $i$ has degree $k$. Its relevance in the number of nodes $N_k$ is weighted with its potential maximum degree, computed as the sum of the row $i$ of the adjacency matrix $A$. The aim of this model is to normalize the nodes degrees according to the potential maximum degree of each node.

\subsection{Binomial replicas}
To make the analysis more robust, we experiment a comparison with a synthetic null model having the exact same architecture as the DBN, created as a Binomial Random Graph (also known as Erd\H{o}s-Renyj model, ER hereafter, \cite{newman2003}). The comparison between the degrees distributions of real network and binomial replicas is displayed in Figure~\ref{fig:dds0.4} in the main text. To generate the random network counterparts $G(N,p)$, the same architecture of the real network is used. The probability of edge existence $p$ is then computed as the ratio between the number of the actual edges (once the network has been pruned) and the total number of connections (given by the architecture), $p = \frac{m}{M}$ \cite{latora2017}. The same probability is used to generate random links between all the four layers of nodes of the random network produced. The resulting replicas are random, but not strictly binomial: the degree distribution inevitably deviates from the Poissonian trend, and this is not (only) due to the fact that the number of nodes is relatively small (3784 nodes), but rather to the network bipartite structure.

\subsection{Correction of degrees distribution}
Due to architectural constraints, it is not possible to observe the idealized power-law and binomial degrees distributions for our networks. Binomial networks have a binomial degrees distribution and the probability mass function is expressed by:

\begin{equation}
	P(k) = \binom{N-1}{k} p^k \, (1 - p)^{N-1-k}
	\label{eq:binomial}
\end{equation}
which means that a given node has $k$ links with probability $p$, out of a total of $N-1$ links it could have if the network was fully connected, i.e. $p = 1$. The number of connections of a given node ranges between virtually zero and its potential maximum degree, which is the total number of neurons of the neighboring layers.

\begin{figure}[t]
    \centering
    \begin{subfigure}[b]{0.3\textwidth}
        \includegraphics[width=1\textwidth]{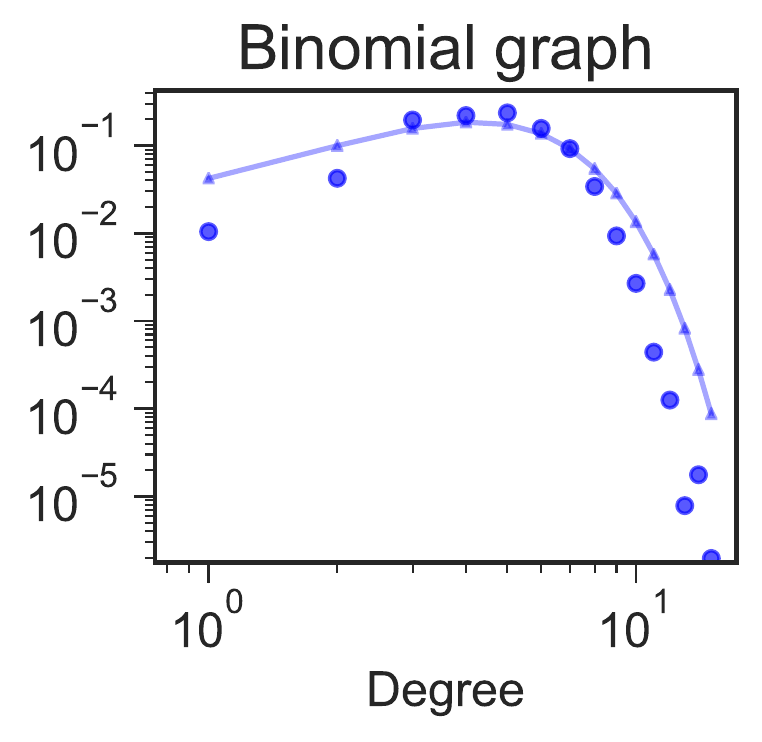}
        \caption{$\tilde{p}$ on $G_{NN}(p)$.}
    \end{subfigure}
    ~
    \begin{subfigure}[b]{0.3\textwidth}
        \includegraphics[width=1\textwidth]{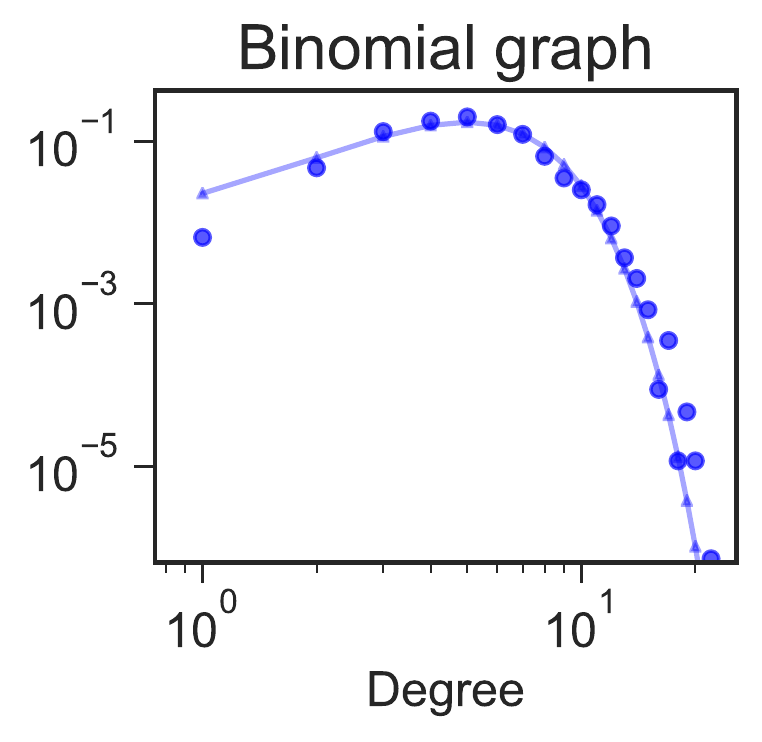}
        \caption{$\tilde{p}$ on $G_{NM}(p)$.}
    \end{subfigure}
    ~
    \begin{subfigure}[b]{0.3\textwidth}
        \includegraphics[width=1\textwidth]{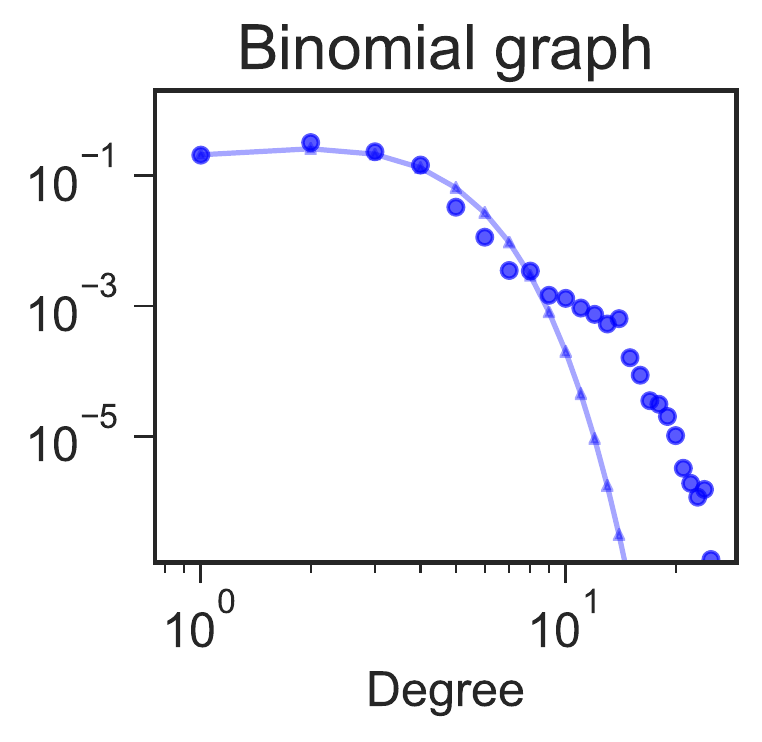}
        \caption{$\tilde{p}$ on $G_N(p)$.}
    \end{subfigure}
    
    \begin{subfigure}[b]{0.3\textwidth}
        \includegraphics[width=1\textwidth]{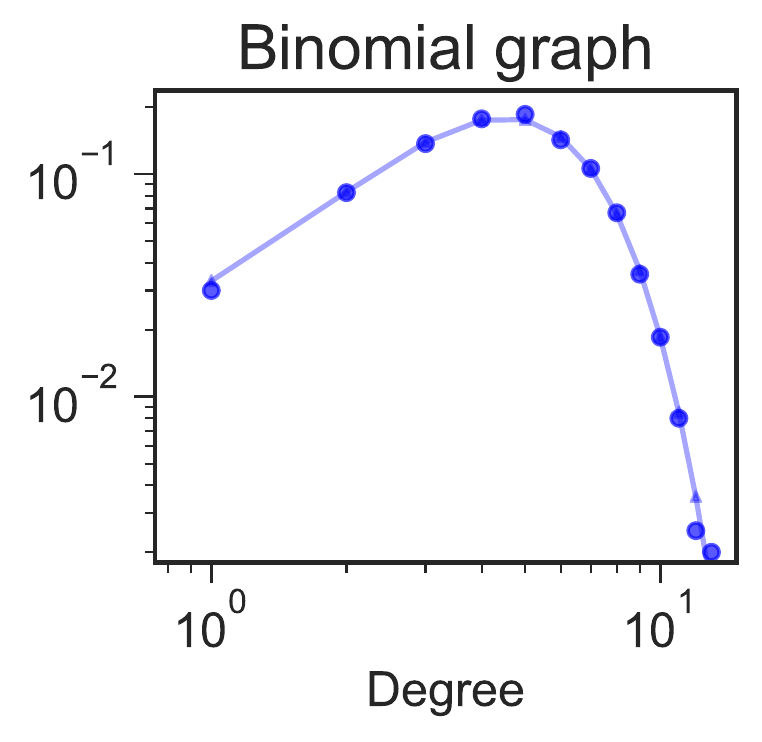}
        \caption{$\tilde{q}$ on $G_{NN}(p)$.}
    \end{subfigure}
    ~
    \begin{subfigure}[b]{0.3\textwidth}
        \includegraphics[width=1\textwidth]{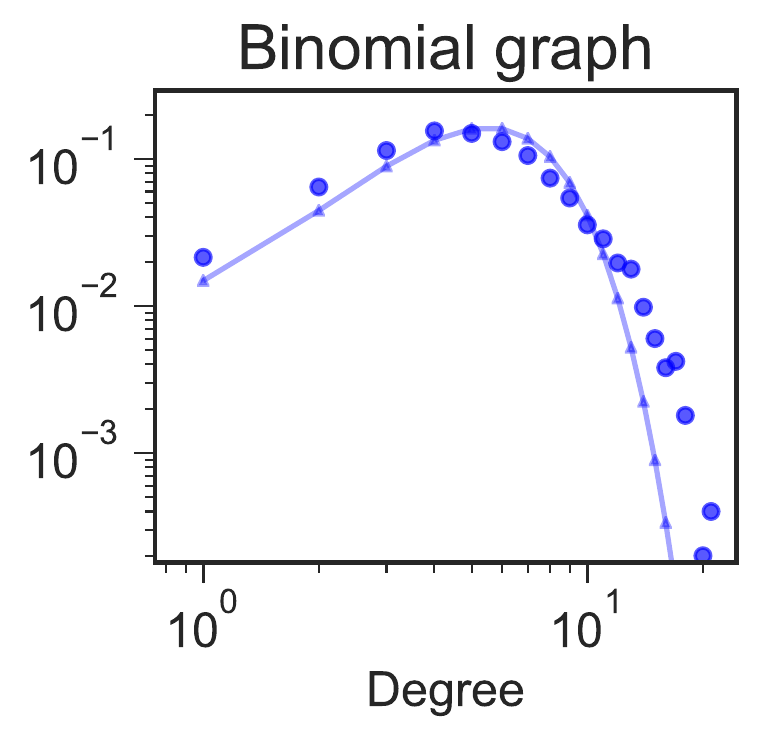}
        \caption{$\tilde{q}$ on $G_{NM}(p)$.}
    \end{subfigure}
    ~
    \begin{subfigure}[b]{0.3\textwidth}
        \includegraphics[width=1\textwidth]{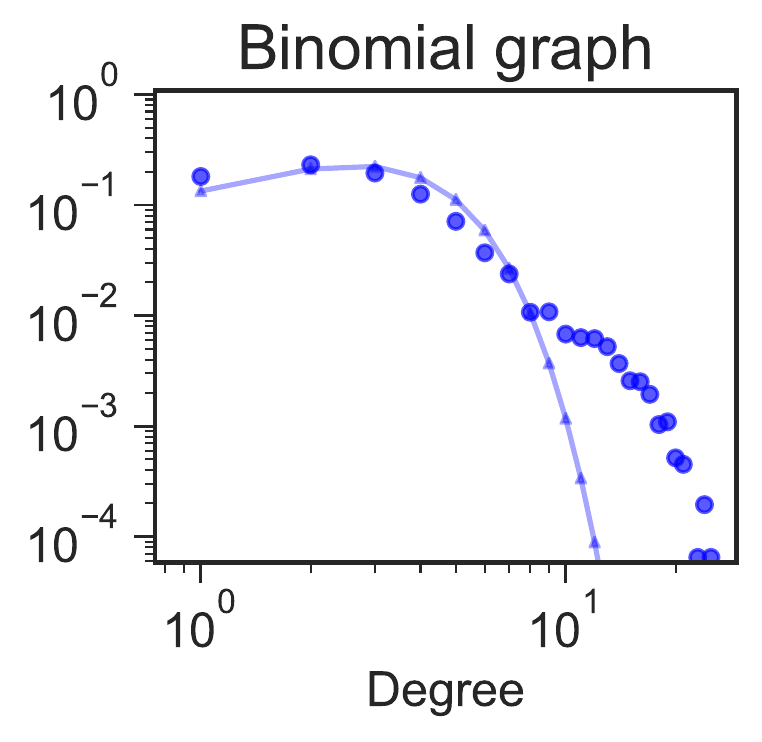}
        \caption{$\tilde{q}$ on $G_N(p)$.}
    \end{subfigure}
    \caption{Experiments on the null models $G_{NN}(p)$, $G_{NM}(p)$ and $G_N(p)$. The value of the probability $p$ is kept to $0.01$, while $N = 1000$ for $G_{NN}(p)$, $M = 2000$ for $G_{NM}(p)$ and $G_N(p)$ retains the structure of the DBN.}
    \label{fig:G_NMp}
\end{figure}

We propose a correction to normalize the node degrees in such a way to correct the architectural bias. The strategy chosen for generating the results in the main text is that of Equation~\ref{eq:ddcorr}, however the choice of this model is not unique. For example, we could choose a model that sets the fraction of nodes having degree $k$ to the corrective factor previously multiplied by $N_k/N$:

\begin{equation}
    \tilde{q}_k = \sum_{i=1}^{N} \mathbb{I}_{k_i = k} w_i = \sum_{i = 1}^{N} \mathbb{I}_{k_i = k} \dfrac{1}{\sum_{j = 1}^{N} A_{ij}}
\end{equation}
again being $A$ the full-graph adjacency matrix. This fraction can be computed easily, then the actual distributions are computed normalizing the raw fractions $\tilde{p}$ and $\tilde{q}$ as follows:

\begin{equation}
    P_k = \frac{\tilde{p}_k}{\sum_k \tilde{p}_k} \qquad Q_k = \frac{\tilde{q}_k}{\sum_k \tilde{q}_k}
\end{equation} 

The two model distributions have been chosen performing some observation on completely random graphs, completely independent of the probabilities of edge existence of the real network. The first synthetic graph $G_N(p)$ shares the same architecture of the DBN, with tunable probability of existence $p$, chosen small. A second test case involves a bipartite random graph with the same number of nodes $N$ on both sides. The third test case involves a bipartite graph having $N$ and $M$ nodes on the two sides. 
The goal is to compare qualitatively both the distributions $P$ and $Q$ on these three test cases to check whether one or the other model distributions perform better. Figure~\ref{fig:G_NMp} displays the results of these simulations. The experiments reveal that $P$ adheres better to the Poissonian trend, thus motivating the choice of the model distribution $P$ for the analyses presented in the main text.

\end{appendices}

\begin{acknowledgements}
We are grateful to Michele De Filippo De Grazia for support with the simulations related to continual learning, and to Ivilin Stoianov for useful discussions about the iterative version of the developmental algorithm presented in this paper.

\vspace{0.2cm}

{\noindent \textbf{Author Contributions} A.T. and M.Zo. contributed to the study conception and design. Computational simulations, data collection and analysis were performed by M.Za. The first draft of the manuscript was written by M.Za. and A.T. All authors commented on previous versions of the manuscript, and all authors read and approved the final manuscript.}

\vspace{0.2cm}

{\noindent \textbf{Compliance with Ethical Standards} Ethical approval: This article does not contain any studies with human participants or animals performed by any of the authors. Funding: This work was supported by a Cariparo Excellence Grant 2017 (project ``Numsense'') to M.Zo.}

\vspace{0.2cm}

{\noindent \textbf{Data Availability Statement} The complete source code of the iterative learning algorithm and additional data related to the analyses discussed in the current study are freely available though the following repository: https://github.com/MatteoZambra/Developmental-Approach-DBN.}

\end{acknowledgements}


\end{document}